\documentclass[lettersize,journal]{IEEEtran}
\usepackage{amsmath,amsfonts}
\usepackage{algorithmic}
\usepackage{algorithm}
\usepackage{array}
\usepackage[caption=false,font=normalsize,labelfont=sf,textfont=sf]{subfig}
\usepackage{textcomp}
\usepackage{stfloats}
\usepackage{url}
\usepackage{verbatim}
\usepackage{graphicx}
\usepackage{multirow}
\usepackage{booktabs}
\usepackage{cite}
\usepackage{lineno,hyperref}
\usepackage{colortbl}
\usepackage{amsmath,amssymb,amsfonts}
\usepackage{algorithmic}
\usepackage{graphicx}
\usepackage{float}

\usepackage[final]{changes} 

\begin{document}

\title{SUES-200: A Multi-height Multi-scene Cross-view Image Benchmark Across Drone and Satellite}

    \author{Runzhe Zhu, Ling Yin, Mingze Yang, Fei Wu, Yuncheng Yang, Wenbo Hu
    \thanks{Runzhe Zhu, Ling Yin, Mingze Yang, Fei Wu, Yuncheng Yang are with the School of Electronic and Electrical Engineer, Shanghai University of Engineering Science, Shanghai 201602, China (email: m025120503@sues.edu.cn; lyin@sues.edu.cn; ymz871500142@163.com; fei\_wu1@163.com; shawn.yangyc@foxmail.com. ) Corresponding author: Fei Wu.}
    \thanks{Wenbo Hu is with the School of Communication and Information Engineering, Shanghai University, Shanghai 200444, China (email: wenbohu@shu.edu.cn. )}
    }


\markboth{IEEE TRANSACTIONS ON CIRCUITS AND SYSTEMS FOR VIDEO TECHNOLOGY}%
{Shell \MakeLowercase{\textit{et al.}}: A Sample Article Using IEEEtran.cls for IEEE Journals}


\maketitle

\begin{abstract}
Cross-view image matching aims to match images of the same target scene acquired from different platforms. With the rapid development of drone technology, \replaced{cross-view matching by neural network models has been a widely accepted choice for drone position or navigation.} {}  However, \deleted{due to} existing public datasets do not include images obtained by drones at different heights, and the types of scenes are relatively homogeneous, which yields issues in assessing a model's capability to adapt to complex and changing scenes. \replaced{In this end, we present a new cross-view dataset called SUES-200 to address these issues.}{} SUES-200 contains \added{24120} images acquired by the drone at four different heights and corresponding satellite view images of the same target scene. To the best of our knowledge, SUES-200 is the first \added{public} dataset that considers the differences generated in aerial photography captured by drones flying at different heights. In addition, we \replaced{developed}{ build} an evaluation for efficient training, testing and evaluation of cross-view matching models, under which we comprehensively analyze the performance of \replaced{nine architectures.} {feature extractors with different CNN architectures.} \replaced{Then, we propose a robust baseline model for use with SUES-200.}{} Experimental results show that SUES-200 can help the model to learn  \replaced{highly discriminative features of the height of the drone.}{ discrimination at different heights.}

\end{abstract}


\section{Introduction}
\IEEEPARstart{C}{ross-view}\ matching\cite{2016Cross} is an essential topic in computer vision research. This technique can be applied in many domains, such as localization, navigation, autonomous driving, and object detection. Satellite and drone platforms are the primary sources of images used as input for the matching. A standard cross-view matching \replaced{system works as follows.}{works as follows:} Given an image to be retrieved in a query dataset from one view, the matching system finds an image under the exact location in a large-scale candidate (gallery) dataset of another view. \replaced{There are two main tasks:} {For cross-view matching under satellite and drone platforms, two main tasks need to be tackled:}

\begin{figure}[ht]
  \begin{center}
  \includegraphics[width=0.5\textwidth]{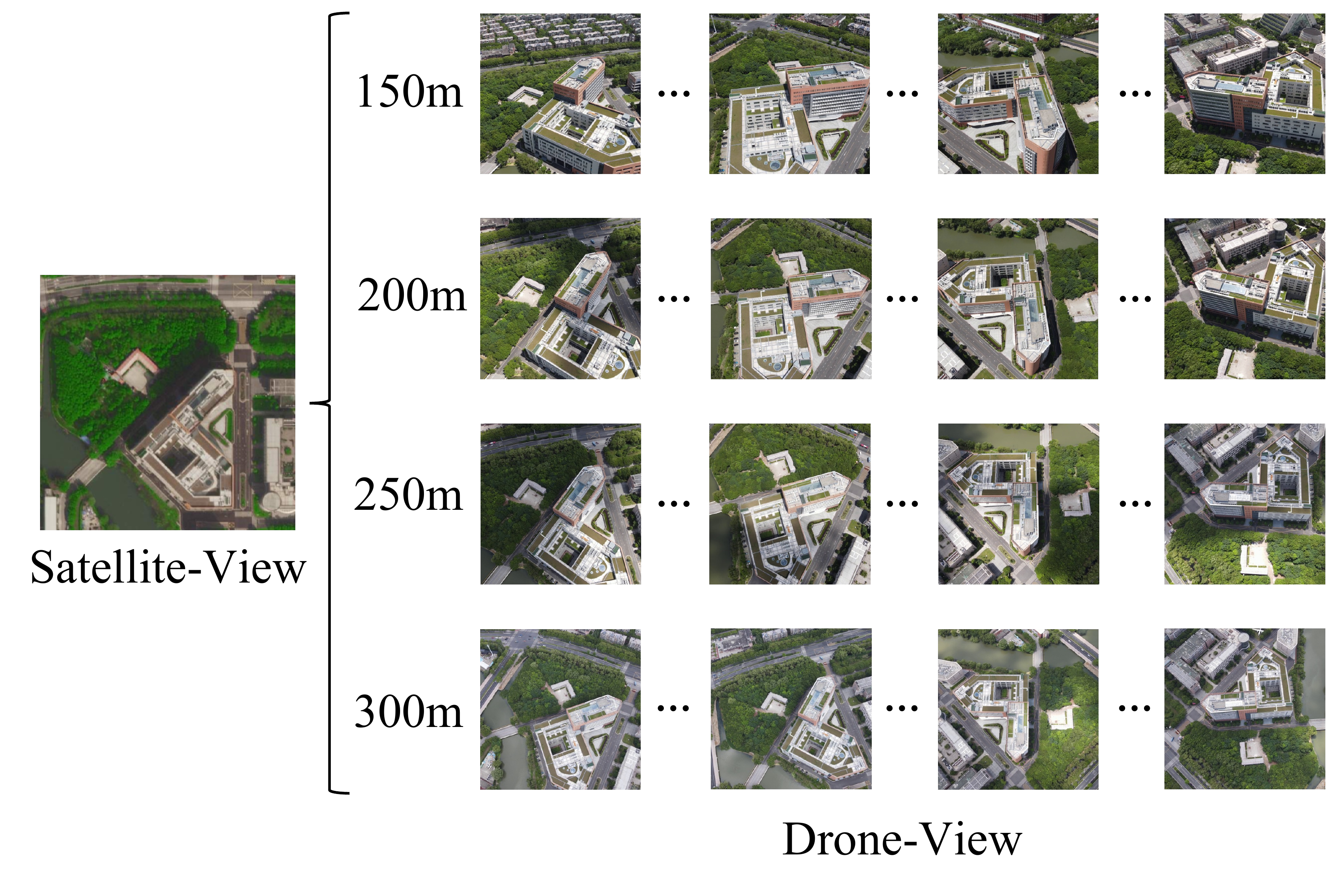}\\
  \caption{A representative target scene of SUES-200 contains fifty drone view images from four heights and one satellite view image.}\label{overview}
  \end{center}
\end{figure}

\replaced{Task 1:}{} Drone \added{view target} localization ($\text{Drone}\rightarrow\text{Satellite}$) and \replaced{Task 2:}{} Drone navigation ($ \text{Satellite}\rightarrow\text{Drone}$). 

Thus, the key to the effectiveness of \added{the matching techniques} is learning \replaced{the discriminative features of images that are invariant under different views}{}.

\deleted{Most of} Previous studies on cross-view matching \cite{7299135,zhai2017predicting,tian2017cross, shi2019spatial,shi2020optimal,shi2020looking} mainly focus on the matching between street view and satellite view, or between street view and bird-view. For example, the datasets CVUSA\cite{workman2015location} and CVACT\cite{liu2019lending} used panoramic street view and satellite views of the same target scene to \replaced{construct}{ form} a cross-view image pair for \replaced{training}{ build} a deep neural network model\deleted{to solve the problem of feature extraction}. \deleted{However,}The quality of matching between street view and satellite images is limited \replaced{by}{to} the much smaller spatial scale of a street view. Thus, \replaced{street view tends to}{easy to} be obscured and interfered with, \replaced{resulting in features not properly extracted by models.} {which leads to the model not extracting the appropriate features.}

With the wide application of drone technology\cite{wang2022auto,zhan2020joint,2022uav}, more and more researchers have been using drone platforms to \replaced{capture}{ describe} target scenes at different spatial and temporal scales. \replaced{Traditionally, image matching of drone view and satellite views}{ The traditional Drone view and satellite view image matching technology} is \replaced{relatively}{ mostly} limited in the military field; fixed-wing drones are conventionally designed to fly at a specific height and collect images in real-time\cite{zhao2019review,zhuo2017automatic,krishnan2021evaluation}. Matching systems are used to match the images captured by a drone with satellite images to infer the drone's location. This autonomous \replaced{locating}{ positioning} system is not affected by the external environment and has strong robustness in complex electromagnetic environments. \added{Recently, rotary-wing drones have been gained wide applications. How to use such vehicles for positioning in low airspace has become a hot research issue. }

Recently, new progress has been made in cross-view view matching research. Zheng et al.\cite{zheng2020university} established the first drone-based multi-source cross-view matching dataset, namely University-1652, which contains images from three perspectives, including street view, aerial drones, and satellite. They also published a baseline for multi-branch CNN networks.\cite{ding2021practical,zhuang2021faster,wang2021each,tian2021uav,dai2021transformer}, and matching accuracy was significantly improved in a more in-depth study. \replaced{However, this dataset still involves a few problems. For example, only synthetic images of drone views are included, which lack realistic variations in lighting. Similarly, differences in images captured by drones at different heights are not distinguished. Moreover, the captured scenes are of a single type, mostly buildings on campuses.}{} \replaced{These problems limit the ability of learning models trained on this dataset to differentiate different types of scenarios. Furthermore, such models are unable to extract robust features from images captured at low heights.}{ These problems lead to poor differentiation of the features extracted by the model and poor robustness of the drone when flying at different heights.}

\replaced{To address these problems}{Therefore}, we propose a multi-height, multi-scene dataset including images from both drones and satellites based on the University-1652 dataset, called SUES-200. \replaced{SUES-200 contains a wider variety of scenes, such as parks, schools, lakes, and public buildings. For each scene, data collected at four different heights(150m, 200m, 250m, and 300m). All of the included images were recorded from onboard drones in flight in real world.}{} SUES-200 contains 200 target scenes, 120 of which are specified for use as a training set, and 80 scenes of which are designated as a testing set. Some samples in SUES-200 are shown in Figure \ref{overview}.

Traditional evaluation metrics for cross-view matching datasets are Recall@K\cite{recall} and AP. \deleted{They cannot fit the new SUES-200 characteristics.} \added{ However, these measures are not suitable for the characteristics of our new SUES-200 dataset, because the differences in drone views at different heights are not taken into account. Moreover, drones encounter diverse interference when flying outdoors.} Therefore, we developed a new evaluation system that focuses on three aspects of the model, including 1) robustness at different heights;  \replaced{2) robustness to uncertainties;}{ preference for two tasks} \replaced{and 3) inference speed;}{ real-time performance during model inference} \deleted{To address the problem that the training and testing process of previous cross-view matching models is complicated, and previous models are not appropriate to modify the model structure.} In addition, we provide a pipeline dedicated to cross-view matching, which helps to improve the efficiency of training, testing, and model evaluation. 

\added{As an experiment, we train and test feature extractors of different deep neural network (DNN) architectures on SUES-200 using the pipeline developed in this work.} The model with the best overall evaluation results is released as the baseline model of SUES-200. We also evaluate the effects of multi-angle feature fusion on matching results and compare the performance of transferred learning models. We perform ablation studies to evaluate each component of the baseline model. Our results show that SUES-200 can help neural models learn high-level features in various scenes captures from different heights. With increasing height, drone footage is gradually less affected by the environment and camera pose and achieves better performance metrics. 

\added{We release a ViT-based model as the baseline of SUES-200. For $ \text{Drone}\rightarrow\text{Satellite}$, baseline achieves 59.32, 62.30, 71.35, and 77.17 Recall@1 accuracy at heights of 150m, 200m, 250m, and 300m, respectively. For $ \text{Satellite}\rightarrow\text{Drone}$, baseline achieves 82.50, 85.00, 88.75, and 96.25 Recall@1 accuracy in 150m, 200m, 250m, and 300m, respectively. This baseline also showed strong robustness to different heights and uncertainties. ViT is a very general scheme compared to other CNN-based algorithms, although its computational complexity is large. The entire dataset, as well as the code for the evaluation, is available at https://github.com/Reza-Zhu/SUES-200-Benchmark.}

The main contributions of this study are summarized as follows.
\begin{itemize}
\item{We build a new cross-view matching dataset: SUES-200, which provides diverse scenes and height views for each scene. All images are acquired in real environments of multiple types of scenes, including real-world light, shadow transformations and disturbances. \added{The datasets, as well as the code for the evaluation, are available at https://github.com/Reza-Zhu/SUES-200-Benchmark.}}
\item{We propose a new evaluation system based on the characteristics of SUES-200 to evaluate the robustness of matching models for different heights, \replaced{robustness to uncertainties}{ the preference for different tasks}, and \replaced{the speed at which inferences are performed}{ and the real-time performance}, together with to the classical Recall@K and AP.}
\item{We establish an efficient pipeline to train and test different matching models and release the baseline model of SUES-200 according to the comprehensive evaluation results.}

\end{itemize}

\section{Related Work}
\subsection{\added{Cross-view Datasets}}
\replaced{Previous cross-view datasets mostly}{ Many previous cross-view datasets} focused on images collected from the same location with different viewpoints via different platforms such as panoramic cameras, satellites, drones, and smartphones. For example, the dataset \cite{7299135} comprises publicly available data, containing a total of 78K data pairs. Each pair consists of two views, including an aerial or bird’s-eye view, and the other view is the street view. Tian \textit{et al.}\cite{tian2017cross} collected images from several locations in a city and constructed image pairs with bird’s-eye and street views. Tian incorporates semantic information to label buildings in images from different views and contains an object detection module in its network structure. The experimental results were evaluated in terms of PR curves and AP. CVUSA \cite{workman2015wide} is a standard cross-view dataset, consisting of image pairs of panoramic street views and satellite views. CVACT\cite{liu2019lending} is a larger panoramic dataset with improved satellite image resolution and more testing sets. Moreover, GPS tags to scenes are supplemented. Both CVACT and CVUSA use Recall@K to evaluate matching results. University-1652 was proposed by Zheng for multi-source cross-view scene matching \textit{et al.} \cite{zheng2020university} as the first geo-localization dataset based on drone footage. It contains image data triplets with satellite, drone, and street views for 1652 buildings in 72 universities. University-1652 generally includes one satellite image, fifty-four images captured by aerial drones, and multiple street view images for a given location. Due to the expense of real-world flight, the drone data in this database was obtained by simulated flights in Google Earth. The drone simulation flight route circles around the target scene and gradually drops in height. University-1652 uses Recall@K and AP to evaluate matching results. Inspired by University-1652, we constructed the SUES-200 dataset to emphasize differences in images acquired by drones at different heights. In addition, we extended the types of scenes, all of which were captured in real scenarios.

\begin{figure}[t]
  \begin{center}
  \includegraphics[width=0.5\textwidth]{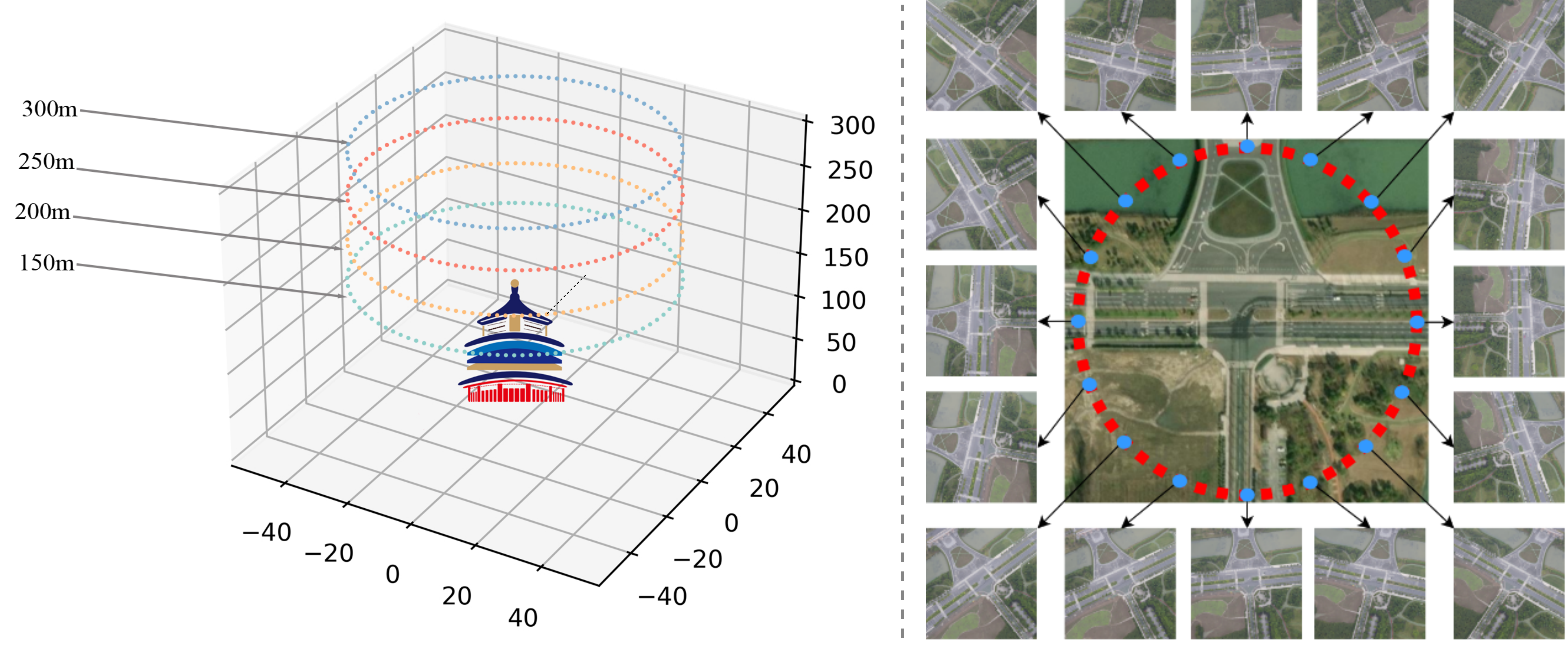}\\
  \caption{The flight height of the drone when collecting images is 150m, 200m, 250m, and 300m. The flight trajectory is one circle around the target scene}\label{fly trace}
  \end{center}
\end{figure}

\begin{table}[t]
\caption{Overview of cross-view approaches.\label{survey}}
\centering
\begin{tabular}{lcc}
\hline
\textbf{Approach} & \textbf{Feature Extractor}  & \textbf{Author}\\
\toprule
CVUSA	         &   		VGG16					&	Workman \textit{et al.}\cite{workman2015wide}     	\\
CVACT	         &   		7-layer CNN				&	Liu \textit{et al.}\cite{liu2019lending}     	\\
University-1652  & 			ResNet-50				&	Zheng \textit{et al.} \cite{zheng2020university}. \\
LCM              &			ResNet-50				&	Ding \textit{et al.}\cite{ding2021practical}	 \\
LPN              &			ResNet-50				&	Wang \textit{et al.}\cite{wang2021each}			\\
PCL              & 			ResNet-50				&	Tian \textit{et al.}\cite{tian2021uav}			\\
MSBA 			 & 			ResNet-50				&	Zhuang \textit{et al.}\cite{zhuang2021faster}			\\
\hline
\end{tabular}
\end{table}

\subsection{\added{Cross-view Methods}}
\added{Traditional cross-view matching methods\cite{lowe2004distinctive,bay2006surf,revaud2019r2d2} are based on hand-crafted feature descriptors such as SIFT\cite{lowe2004distinctive}, SURF\cite{bay2006surf}, and ORB\cite{rublee2011orb}. However, these feature extraction methods are not robust and are susceptible to uncertainties such as lighting and occlusion, especially for drones flying at heights. False or missing matches typically occur frequently due to excessive differences between the acquired images and the satellite view images.} Since the publication of the University-1652 dataset, considerable progress has been made in the past years in deep learning methods. Liu \textit{et al.}\cite{ding2021practical} proposed LCM, which utilized ResNet\cite{he2016deep} as a backbone network and trained the image retrieval problem as a classification problem. The LCM improved the Recall@1 and AP by 5-10 \% over the baseline of University-1652. Wang \textit{et al.}\cite{wang2021each} designed LPN to consider the contextual information of neighboring regions. The LPN used a square-ring partition strategy to divide feature maps, which provided good robustness to changes in rotation. LPN achieved good performance on University-1652, CVACT, and CVUSA. Tian \textit{et al.}\cite{tian2021uav} presented a method that integrated the spatial correspondence between the satellite views and information on the surrounding area. This was performed in two steps, first converting the tilted view of the drone into a vertical view by perspective transformation and then transforming the image of the drone view to be closer to the satellite view using a conditional GAN\cite{mirza2014conditional}. The experimental results show that this method improved accuracy by 5\% over LPN on University-1652. Inspired by the development of attention mechanisms, Zhuang \textit{et al.} \cite{zhuang2021faster} developed MSBA to eliminate the differences in images acquired from different viewpoints. MSBA cuts an image into several parts with different scales. Based on that division, a self-attention mechanism is used for more effective feature extraction. They showed that MSBA performed better than LPN in terms of accuracy and inference efficiency. \added{Table \ref{survey} gives an overview of existing approaches. Importantly, most approaches adopt the same backbone network and were not tested for other feature extractors. In contrast, in the present work, we trained several cross-view matching models and then tested and evaluated the performance of different backbone networks such as VGG\cite{simonyan2014very}, ResNet\cite{he2016deep}, and DenseNet\cite{huang2017densely} in extracting features at different heights using the pipeline.}

\begin{figure}[t]
  \begin{center}
  \includegraphics[width=0.4\textwidth]{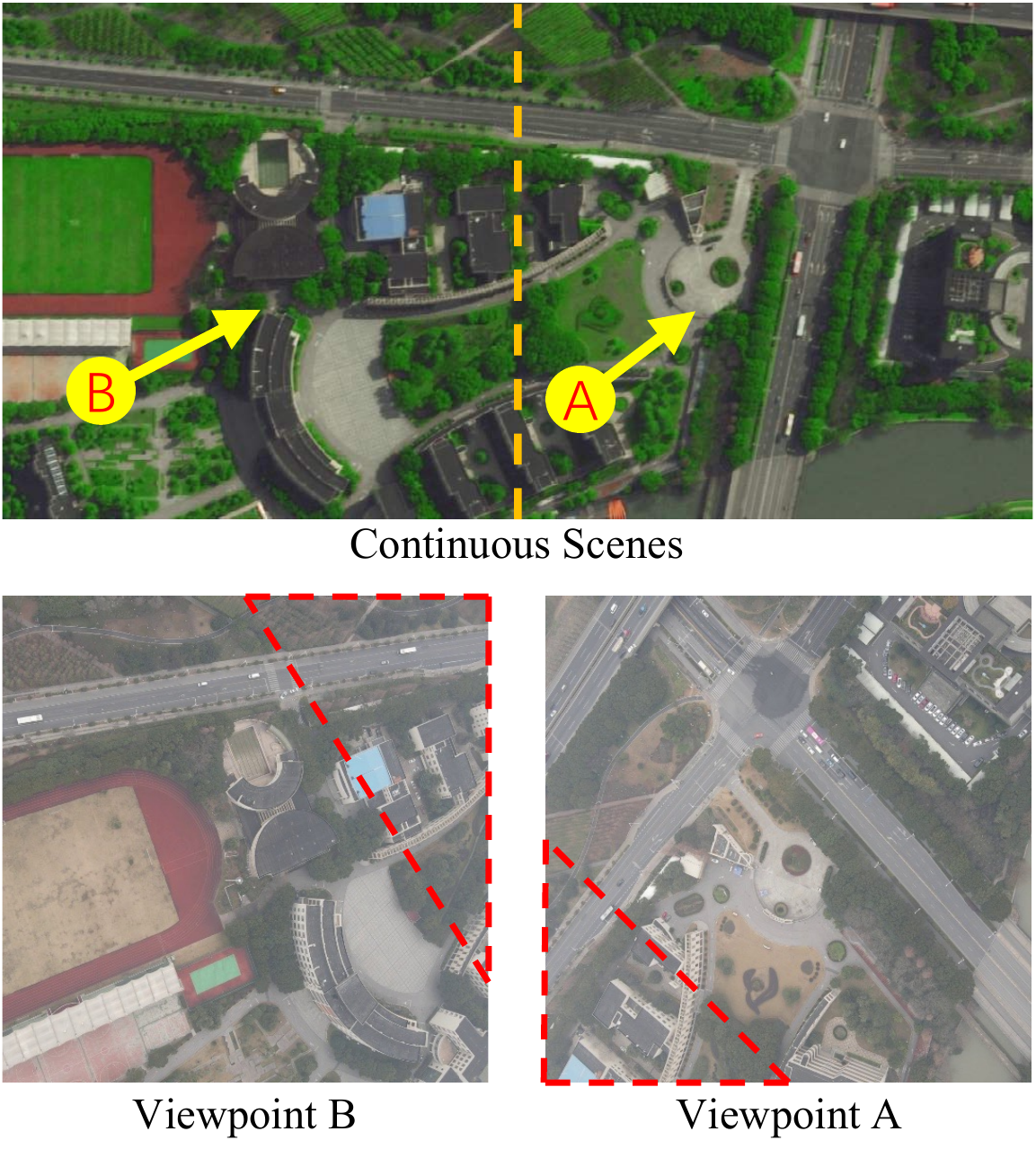}\\
  \caption{\replaced{Continuous Scenes display two continuous satellite view images. Yellow arrows indicate the directions of the drones' viewpoints. Two corresponding drone view images are shown below, with the area overlapping in the images marked by the dashed red line.}{} }\label{similar}
  \end{center}
\end{figure}

\section{SUES-200 Dataset}
\subsection{Dataset Description}
\replaced{SUES-200 is a cross-view matching dataset with the characteristics of multiple sources}{ The cross-view matching dataset has the characteristics of multiple sources}, multiple scenes, and panoramic views. We collected multi-source images of satellite views and corresponding drone views at 200 locations \replaced{around the Shanghai University of Engineering and Science(SUES)}{ in the vicinity of our school}. \added{We used 0001, 0002, ... 0200 to distinguish the images obtained in different scenes, and numbers 1-200 represent specific scenes.} Specifically, to enable the model to learn highly discriminative features at different heights, we collected drone view images at 150m, 200m, 250m, and 300m. \replaced{SUES-200 includes a broader range of scene types, not limited to campus buildings, containing parks, schools, lakes, and public buildings. The rich multi-type scenes enable the models to learn features that can be adapted to real environments.}{}

\begin{table*}[!t]
\caption{Comparison between SUES-200 and other cross-view datasets.\label{datasets}}
\centering
\begin{tabular}{lcccc}
\hline
\multicolumn{1}{c}{Datesets} & SUES-200                                                & University-1652\cite{zheng2020university}        & CVUSA\cite{workman2015wide}            & Tian et al\cite{tian2017cross}.       \\
\hline
\toprule
Platform                     & Drone, Satellite                                         & Drone, Ground, Satellite & Ground, Satellite & Ground, 45° Aerial \\
Target                       & Diversity                                                 & Building               & User             & User              \\
Height difference            & TRUE                                                    & FALSE                  & FALSE            & FALSE             \\
Training                     & 120 * 51                                                & 701 * 71.64            & 35.5k * 2        & 15.7k * 2         \\
Images/Location              & 50 + 1                                                  & 51 + 16.64 + 1         & 1 + 1            & 1 + 1             \\
Evaluation                   & Recall@K \& AP \& Robustness \& Inference Speed & Recall@K \& AP         & Recall@K         & PR\&AP             \\
\hline
\end{tabular}
\end{table*}

\begin{table}[t]
\caption{statistics of SUES-200 training and test sets, including the image number and the scene number of training set, testing set.\label{statistics}}
\centering
\begin{tabular}{lccc}
\hline
\multicolumn{4}{c}{Training Dataset}                          \\
\multicolumn{1}{c}{Views} & Locations & Images at Each Height & Total \\
\hline
\toprule
Drone                     & 120       & 6000            & 24000 \\
Satellite                 & 120       & \_             & 120   \\
\hline
\multicolumn{4}{c}{Testing Dataset}                             \\
\multicolumn{1}{c}{Views} & Locations & Images at Each Height & Total \\
\hline
\toprule
Drone query               & 80        & 4000            & 16000 \\
Satellite query           & 80        & \_              & 80    \\
Drone gallery             & 200       & 10000           & 40000 \\
Satellite gallery         & 200       & \_             & 200  \\
\hline
\end{tabular}
\end{table}

\replaced{Satellite-view images are obtained from AutoNavi Map and Bing Maps. A single satellite image is included for each location. The schematic diagram of the drone flight is shown in Figure \ref{fly trace}. The drone flight path was set to a curve in different heights to capture multi-angle information of target scenes. We sampled 50 frames uniformly from the flight video recorded by the drone. Overall, every location includes one satellite view image and 50 drone view images.}{}

\added{
In addition, multiple satellite images are consecutively selected in the same area by SUES-200. When the drone flies in one of the locations, the image includes information about nearby locations. As shown in Figure \ref{similar}, there is some overlap between drone maps of different scenarios. It is desirable that the cross-view matching models could pay attention to the main feature in the scene without the effect of overlapping regions.
}

In order to prevent information loss due to image resolution, \replaced{drone images in SUES-200 use the original resolution of $\text{1080} \times \text{1080}$ and satellite images use the resolution of $\text{512} \times \text{512}$}{ both drone images and satellite images in SUES-200 use the original resolution of $\text{1080} \times \text{1080}$ and $\text{512} \times \text{512}$.}. The dataset includes 200 locations with 50 drone images and 1 corresponding satellite image for each location. SUES-200 is divided into training and testing sets, \replaced{with 120 locations designated for training and 80 locations as testing data.}{ where 60\% is training data and 40\% is test data. } To accomplish the two tasks mentioned in the introduction, the testing data include the query drone dataset, query satellite dataset, gallery drone dataset, and gallery satellite dataset. Among these, the gallery dataset contains the testing data and adds the training data as confusion data to increase the difficulty of matching. 

\added{In the testing phase, we consider Task 1 and Task 2 as image retrieval tasks. Taking $\text{Drone}\rightarrow\text{Satellite}$ as an example, the query is a drone image, the gallery is satellite image. The model first extracts features from the images in the gallery set and stores them locally. Then, a single query image is fed into the model to extract features and calculate the distance between the query feature and gallery images. The image pair with the shortest distance between the drone view image and the satellite view image is considered the matching result.} Some statistics on the datasets are shown in Table \ref{datasets} and Table \ref{statistics}.

Finally, we summarize the new characteristics of the SUES-200 dataset.
\begin{enumerate}
\item{\textbf{Multi-height:} SUES-200 contains data collected at different heights: 150m, 200m, 250m, and 300m, and can evaluate model metrics at different heights. To the best of our knowledge, SUES-200 is the first cross-view dataset to include images recorded by cameras on drone vehicles flying at different heights.}
\item{\textbf{Multi-scene:} SUES-200 contains data from different types of scenes. This can help models extract invariant features in more scenes and expand the scope of scene applications for drone-based cross-view matching techniques.}
\item{\textbf{Continuous-scenes:} Some of the included target scenes were collected in the same area. \replaced{The drone image is thus affected by the surrounding scene; for example, information from another scene may be recorded. This poses a challenge to the ability of trained models to differentiate between scenes, but this is also realistic for practical application environments.}{}}
\end{enumerate}

\begin{figure}[t]
  \begin{center}
  \includegraphics[width=0.45\textwidth]{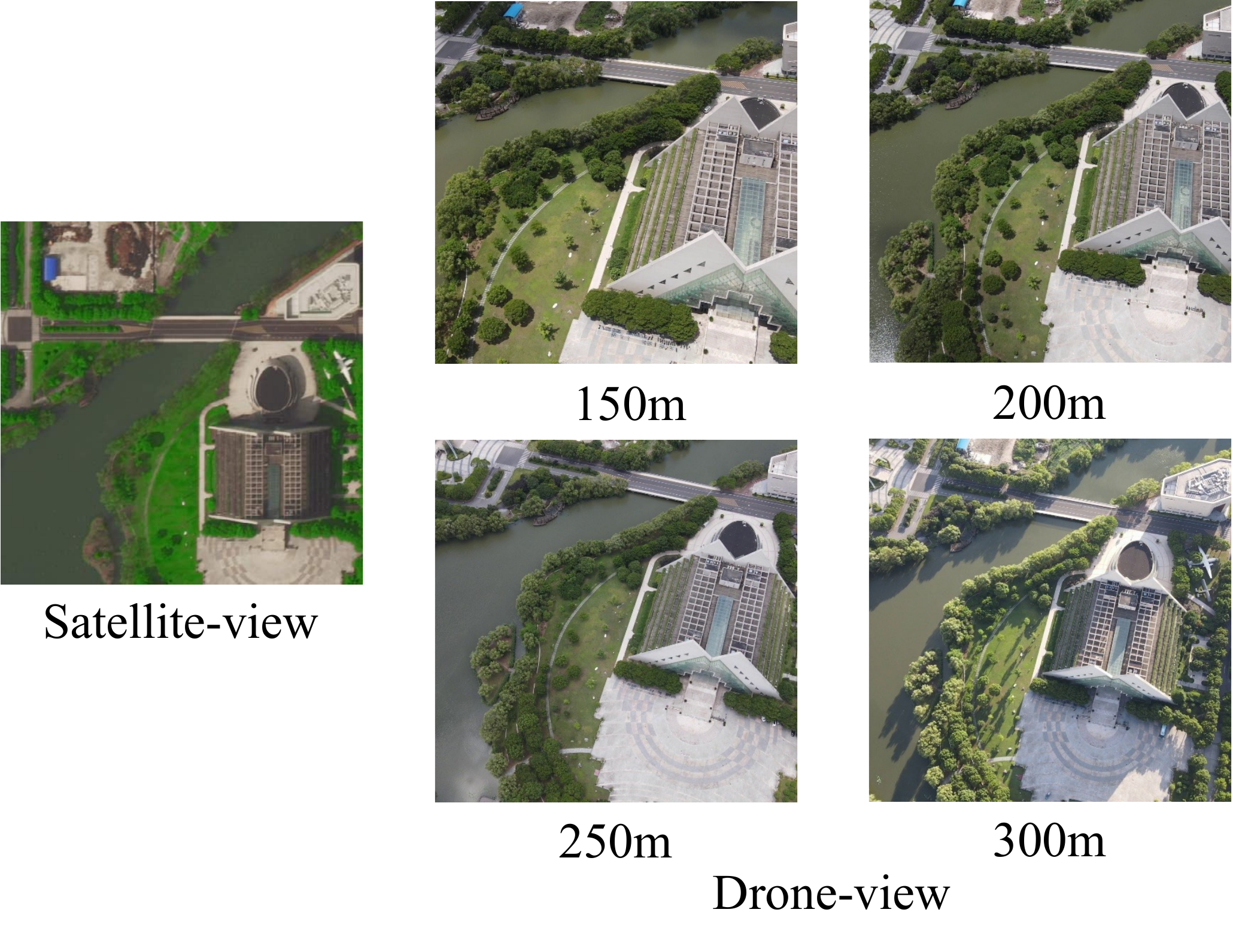}\\
  \caption{As the height rises, the images captured from the perspective of the drone become increasingly similar to the satellite view.}\label{Robustness_Height}
  \end{center}
\end{figure}

\subsection{Evaluation Protocol}
In this subsection, we introduce the evaluation system of SUES-200. In response to the existing real-world problems, in addition to the traditional Recall@K\cite{liu2019lending,vo2016localizing,zhai2017predicting} and AP\cite{lin2015learning,tian2017cross} evaluation metrics, we propose a method to measure model robustness at different heights, as well as a method to measure the robustness of trained models to uncertainties and a method to evaluate inference speed.

\textbf{Recall@K and AP.}  SUES-200 contains 200 target scenes, including 120 scenes for training and 80 scenes for testing. Among these, 120 scenes from the training set are also included in the gallery as distractors. There is no overlap between the training and testing data. \replaced{Recall@K (R@K) represents the probability that a correct match appears in the top-k ranked retrieval results. Recall@1 is very sensitive to the position of the first true-matched image appearing in the ranking of the matching result. A higher recall score shows a better performance of the network.}{ Recall@K is very sensitive to the position of the first true-matched image appearing in the ranking of the matching result. Therefore, it is suitable for a test dataset that contains only one true-matched image in the candidate gallery. } The AP is the area under the precision-recall(PR) curve, which considers the position of all true-matched images in the evaluation. Recall@K is defined as follows.

\begin{equation}
\text{Recall@K} = \begin{cases}
          \ 1, \quad  if \  order_{true} < K + 1\\
          \ 0,  \quad    otherwise \\
\end{cases}
\end{equation}

AP is formulated as follows:
\begin{equation}
        \text{AP} = \frac{1}{m} \sum_{h=1}^{m}\frac{p_{h-1}+p_h}{2} ,
        \text{where} \ p_0 =1 \\ 
\end{equation}

\begin{equation}
	 p_h = \frac{T_h+1}{T_h+F_h}
\end{equation}

where $m$ is the number of true-matched images for a query, $T_h$ and $F_h$ are the numbers of true-matched images and false-matched images before the $(i + 1)$-th true-matched image in the matching.

\textbf{Robustness at different heights.}  SUES-200 differentiates the images acquired by drones at different heights, as shown in Figure \ref{Robustness_Height}. Measuring the robustness of the model at different heights is also an important evaluation index. \deleted{We present a method to evaluate the model's robustness of flight heights as follows.} \added{The drone images appear most similar to satellite view images at 300m. As the height decreases, the drone's field of view gradually narrows. The size of the target scene becomes larger, and more detailed information is presented. These factors make it increasingly difficult for the model to distinguish the variations across scenarios. To evaluate the model's robustness to height, we set the Recall@1 at 300m as the baseline. Then, the Recall@1 at other heights is divided by the baseline to evaluate the reduction in accuracy with height, which is calculated as follows.}

\added{
\begin{equation}
RDR_{n} = 1 - \frac{R@1_{n}}{R@1_{300m}} \\
\end{equation}
}

\added{
$RDR_{n}$ represents the recall degradation rate (RDR) of the model at a given height $n$. The RDR of the model between 300m to 150m directly reflects the robustness of the height of the model. Finally, we denote the overall robustness protocol of the height as:}
\added{
\begin{equation}
RDR_{150m} = 1 - \frac{R@1_{150m}}{R@1_{300m}} \\
\end{equation}
}

\added{
The larger the $RDR_{150m}$, the less robust the model is to height, and vice versa.
}

\begin{figure}[t]
  \begin{center}
  \includegraphics[width=0.45\textwidth]{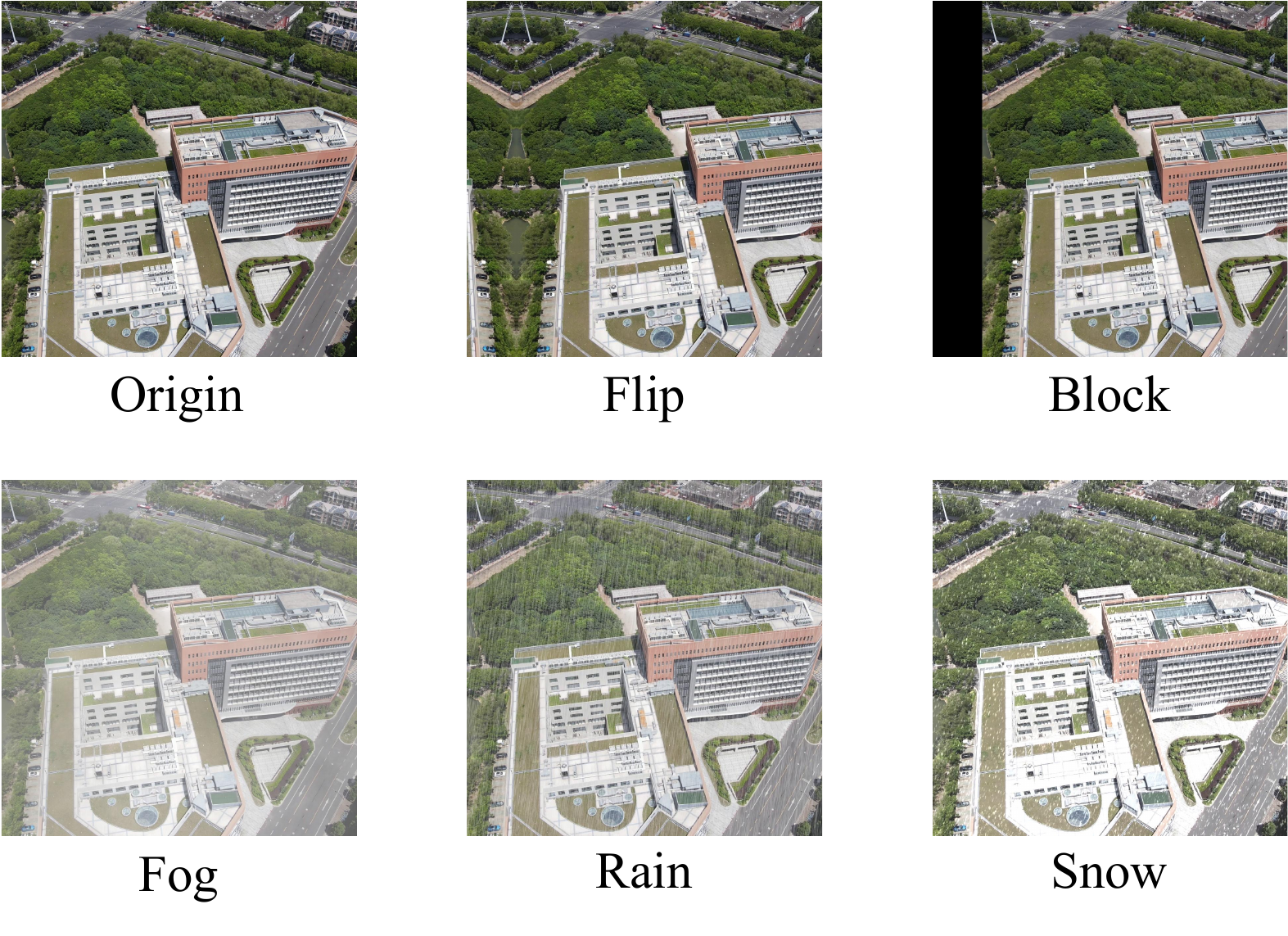}\\
  \caption{We list five uncertainties including flip, block, fog, rain, and snow.}\label{Robustness_Factor}
  \end{center}
\end{figure}

\added{
\textbf{Robustness to uncertainties.}
In practice, the images captured by drone vehicles are often disturbed by various uncertainties, such as the target being obscured or offset and various weather factors. To evaluate the performance of the model under those uncertainties, we simulated these effects by applying augmentation to the drone images in the query set. We considered five types of factors, including flip, block, fog, snow, and rain, in Figure \ref{Robustness_Factor}. We denote the original AP of the model as $AP_{origin}$, and the disturbed AP at a certain height as $AP_{i}$. The $n=4$ indicates four times this height. We calculate the average rate of degradation of precision (RDP) for a given model at four heights to indicate its robustness. The equation is shown as follows.}
\added{
\begin{equation}
RDP =  \frac{\sum\limits^{n=4}_{i=1}1 - AP_{i}/AP_{origin}}{4} \\
\end{equation}
}

\replaced{\textbf{Inference Speed.}}{\textbf{Real-time.}}In the actual application process, the inference speed of the model is a significant concern. Therefore, we refer to the formula mode of \cite{zhuang2021faster} to evaluate the inference speed. \cite{zhuang2021faster} proposed a "real-time" method to evaluate the inference time of a single query.  \added{However, we considered that this approach is not sufficiently comprehensive to evaluate the overall performance because its definition of "real-time" only focuses on the inference time of the query image. Therefore, we present inference speed to assess the combined inference time of the query and gallery images. } We chose a base model with relatively fast inference performance and take its inference time as the benchmark speed. The benchmark is set as $1.00$, and other inference speeds are then denoted as $1.00 \times T (0 < T < +\infty)$.

\begin{figure*}[t]
  \begin{center}
  \includegraphics[width=\textwidth]{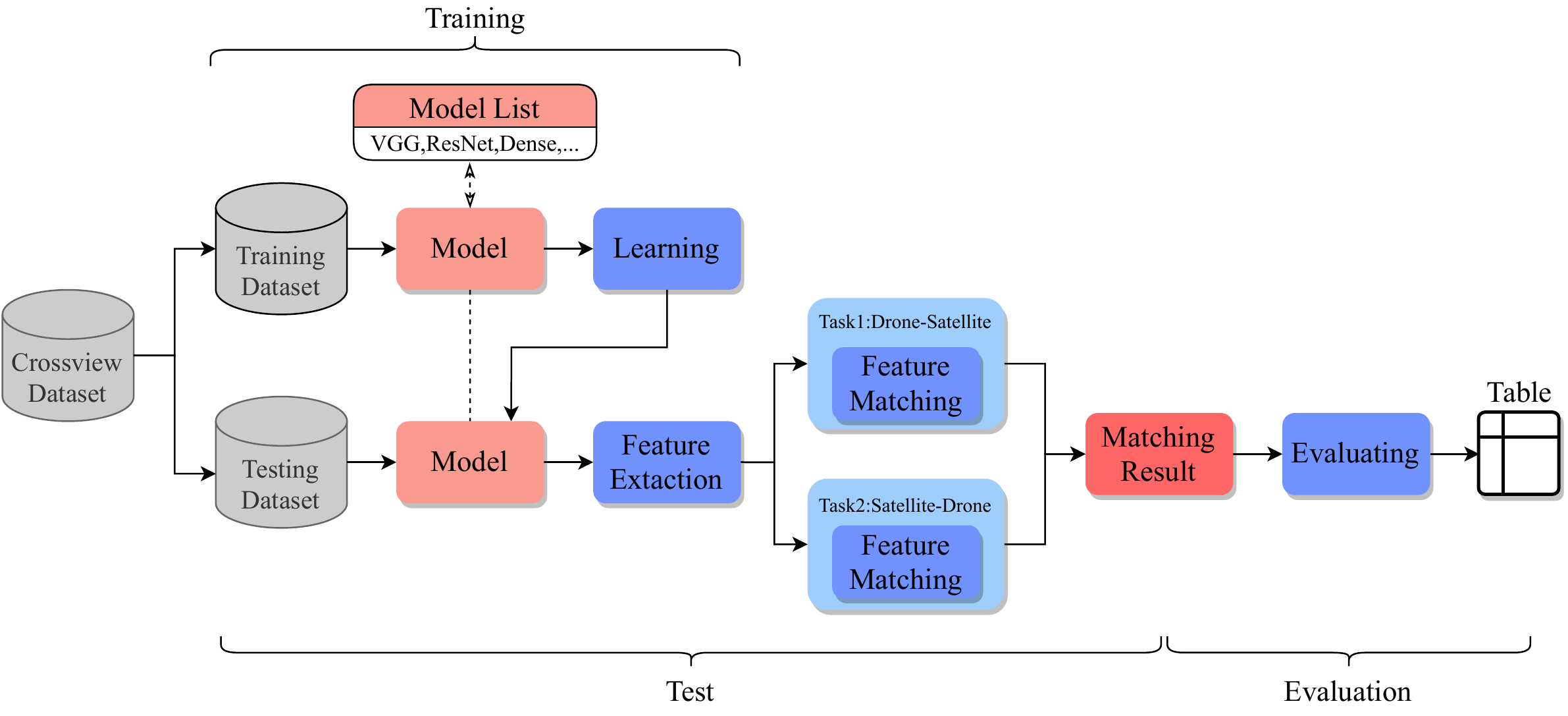}\\
  \caption{The pipeline reads images from the dataset and sends them to the currently selected model for training. After training, the model with the best parameters is selected and sent to Task 1 or Task 2 for testing, and the evaluation module evaluates the test results to form an evaluation table.}\label{pipeline}
  \end{center}
\end{figure*}

\section{Method}
\subsection{Pipeline}
We established a pipeline to solve the cross-view matching problem. This approach provides a standardized method to efficiently train, test, and evaluate different models. As shown in Figure \ref{pipeline}, in this pipeline, the input is the cross-view matching dataset, and the output is the values of each evaluation index, in which the model is a deep neural network constructed by the user. The network is divided into a backbone network and a classification network part. Selecting different feature extractors in the “Model List” replaces the backbone network part in the corresponding network structure, and the user can also customize its network structure. Details of the network structure of deep neural networks are provided in the next section. The images in the testing set are input to the model, which extracts features and completes Task 1: $ \text{Drone}\rightarrow\text{Satellite}$, and Task 2: $\text{Satellite}\rightarrow\text{Drone}$, and the obtained feature matching results are finally passed to the evaluation unit to obtain the evaluation table.
\subsection{Network Architecture and Loss Function}
\begin{figure*}[t]
  \begin{center}
  \includegraphics[width=\textwidth]{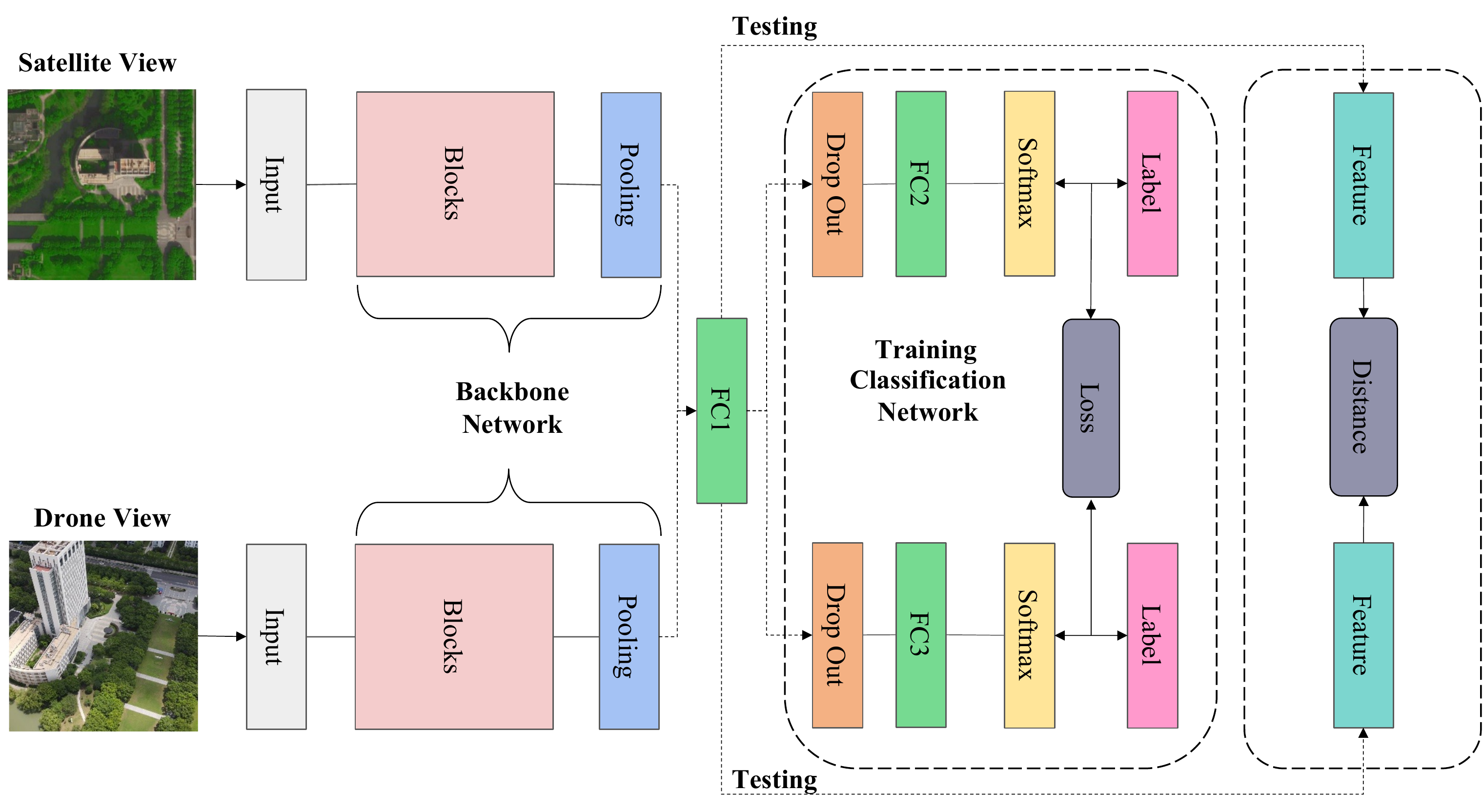}\\
  \caption{Basic network architectures for cross-view matching. We apply two-branch network structures with cross-entropy loss to train the model. \added{The feature map is extracted by the backbone network and pooling layer. Then, a shared weight FC layer(FC1) unifies the feature dimension. Next, the feature map is fed into the classifier network for training.} In addition, the cosine distance is used to calculate the similarity between the query and candidate images in the gallery for testing.}\label{network}
  \end{center}
\end{figure*}

The drone and satellite images included in SUES-200 originate from different sources, but there are still some similarities. Our \deleted{Our goal in designing the}deep learning network extracts robust and invariant features separately and maps them to a high-dimensional space for the following matching task. After referring to previous studies, we constructed a two-branch deep \deleted{convolutional} neural network\added{(DNN)} architecture. One branch \deleted{is used to} extracts feature from satellite view images, and the other branch \deleted{is used to} extracts features from drone view images. To test the performance of different DNN structures on different source image feature extractors,  we apply network structures that extract features in two branches of backbone networks that are replaceable. \added{Subsequently, we add a shared weight fully connected(FC) layer to unify the feature dimensions.} In the training process, \added{we add an MLP block, including a drop-out layer, an FC layer, and a softmax layer, at the end of the branch to treat the processing as a classification task.} Each target location is treated as a class to train the entire network. \added{In the testing process, the feature map is unified by FC1, and then the distance between each feature is calculated by distance measurement algorithms.} The architecture of the network is shown in Figure \ref{network}.

In recent years, different \replaced{DNN}{CNN} structures have been \replaced{extensively}{greatly} developed. ResNet\cite{he2016deep} is widely used as a CNN-based backbone network\cite{zhuang2021faster,tian2021uav,ding2021practical,wang2021each} for feature extraction in the field of cross-view matching due to its clever design structure and excellent performance.  With further research on ResNet and the emergence of attention mechanism, ResNet has been further improved to produce extended variants such as SE-ResNet\cite{hu2018squeeze}, ResNeSt\cite{zhang2020resnest}, CMAB-ResNet\cite{woo2018cbam}, and these models have achieved excellent performance on image classification datasets such as ImageNet\cite{deng2009imagenet}. In addition to ResNet, other structured CNNs are also a topic of considerable research interest in recent years, including DenseNet\cite{huang2017densely}, EfficientNet\cite{tan2019efficientnet}, Inception\cite{szegedy2017inception}. \added{Moreover, ViT\cite{dosovitskiy2020image} architecture has achieved great success in various computer vision tasks. Is there a more proper feature extractor than ResNet in cross-view matching?} In our experiment, we tested the improved CNN-based and other structures on SUES-200 and evaluated these models according to the evaluation system.

For the loss function, \replaced{since}{ because} the model training process is considered a multi-classification task, we adopted a cross-entropy as the loss function, \deleted{which is typical in multi-classification tasks}. Cross entropy is mainly used to determine how close the actual output is to the expected output, i.e., the smaller the cross entropy between the network output and the labels, the better the classification ability of the network. $z^i_j(y)$ is the logarithm of $ \text{ground-truth} \ y$, and  $\hat{p}(y|x^i_j)$ is the probability of the predicted outcome of the model equal to  $\text{ground-truth} \ y$. The mathematical formula is given as follows.

\begin{equation}
\hat{p}(y|x^{i}_j) = \frac{exp(z^{i}_j(y))}{\sum^{C}_{c=1}exp(z^i_j(c))}
\end{equation}

\begin{equation}
\text{Loss}=\sum\limits_{i,j}-log(\hat{p}(y|x^i_j))
\end{equation}

In the two-branch DNN, both outputs of the model need to be compared with the label to obtain two loss values. Let the loss of drone view be $L_d$, and the loss of satellite view be $L_s$, and these two loss values are added to get $L_{total}$. We optimize the whole network through $L_{total}$. 

\begin{equation}
L_{total} = L_s + L_d,
\end{equation}

the query images in the test set are from a drone view and a satellite view. We feed the query images to the model with fixed parameters, remove the classification network from the training layer, and use the backbone network to output the feature vectors directly. The feature vector of the drone-view image is represented as $f_d$, and the feature vector of the satellite view is defined as $f_s$. Our test aim is to find the most similar set of feature vectors by cosine distance to measure the similarity between $f_d$ and $f_s$. $f_di$ and $f_si$ are parts of the feature vector, and a smaller cosine distance means that the set of features is less similar. A larger cosine distance implies that this pair of feature maps are more similar to each other. The formula is given as follows.

\begin{equation}
\text{Cosine} = \frac{f_df_s}{||f_d||\times ||f_s||} =  \frac{\sum\limits_{i=1}^{n}f_{di}f_{si}}{\sqrt{\sum\limits_{i=1}^{n}{(f^{2}_{di})}}\sqrt{\sum\limits_{i=1}^{n}{(f^{2}_{si})}}}
\end{equation}

\begin{figure*}[t]
\centering
\begin{minipage}[t]{0.24\linewidth}
\centering
\centerline{\includegraphics[width=1.3\textwidth]{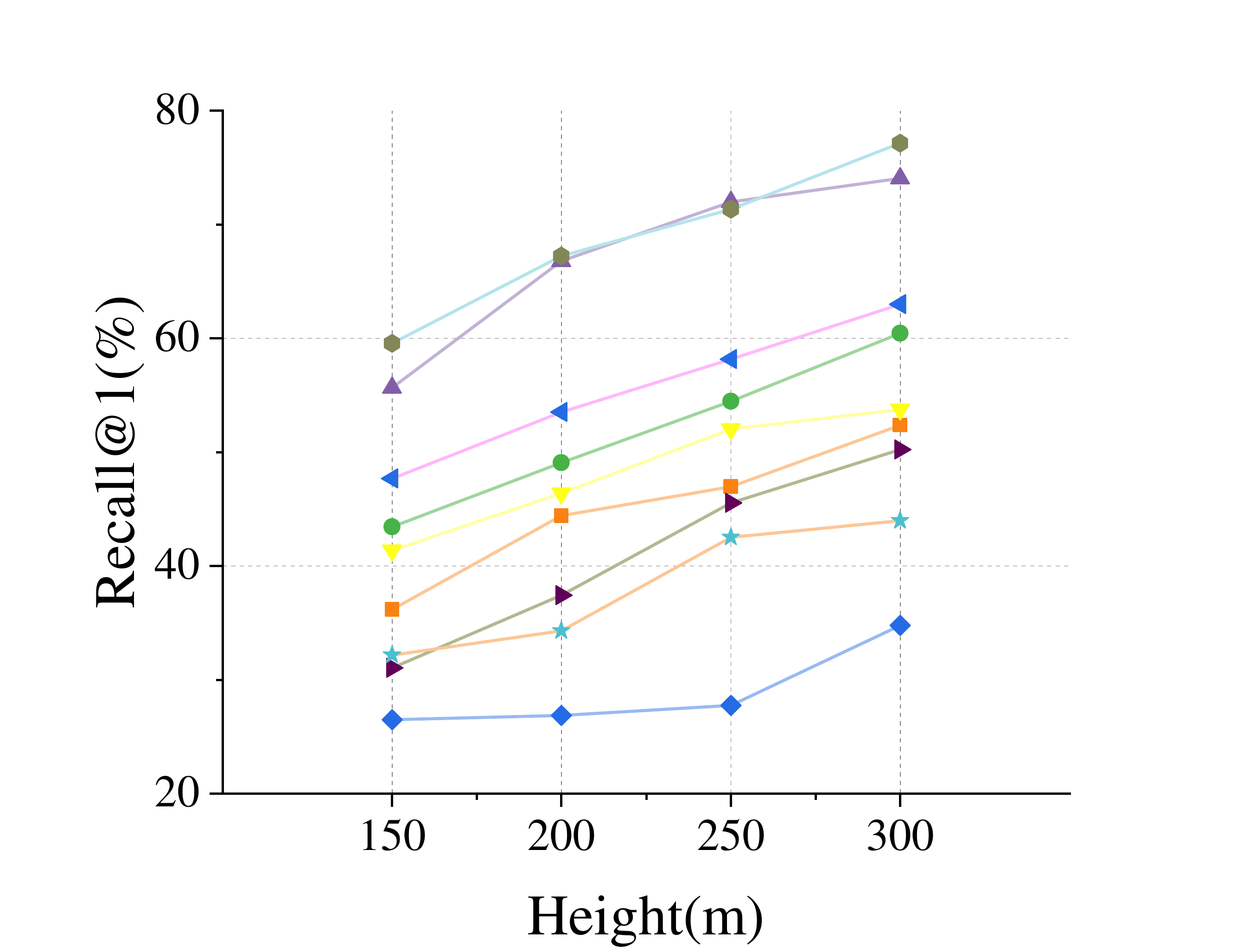}}
\centerline{(a)}
\end{minipage}%
\centering
\begin{minipage}[t]{0.25\linewidth}
\centerline{\includegraphics[width=1.3\textwidth]{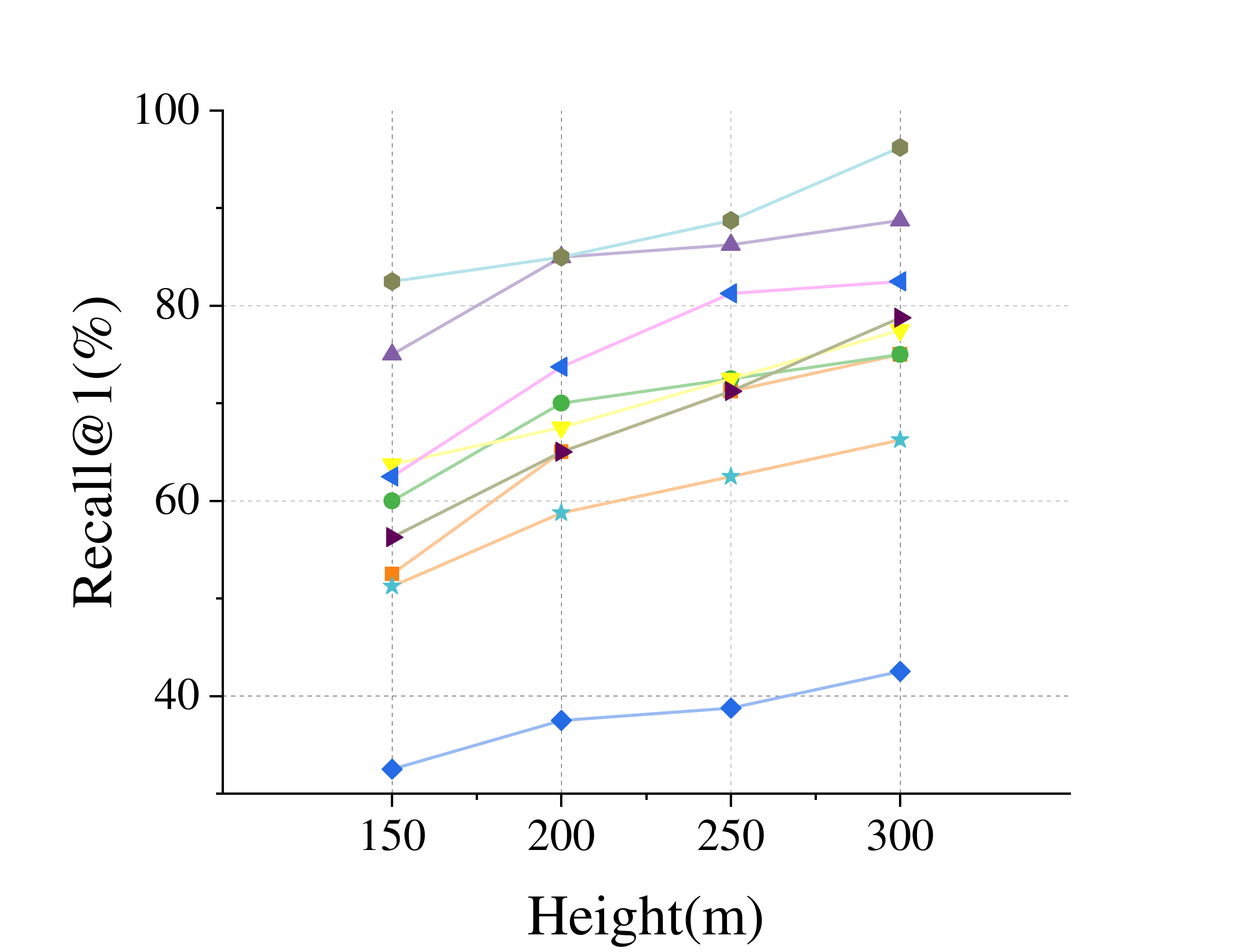}}
\centerline{(b)}
\end{minipage}%
\centering
\begin{minipage}[t]{0.25\linewidth}
\centerline{\includegraphics[width=1.3\textwidth]{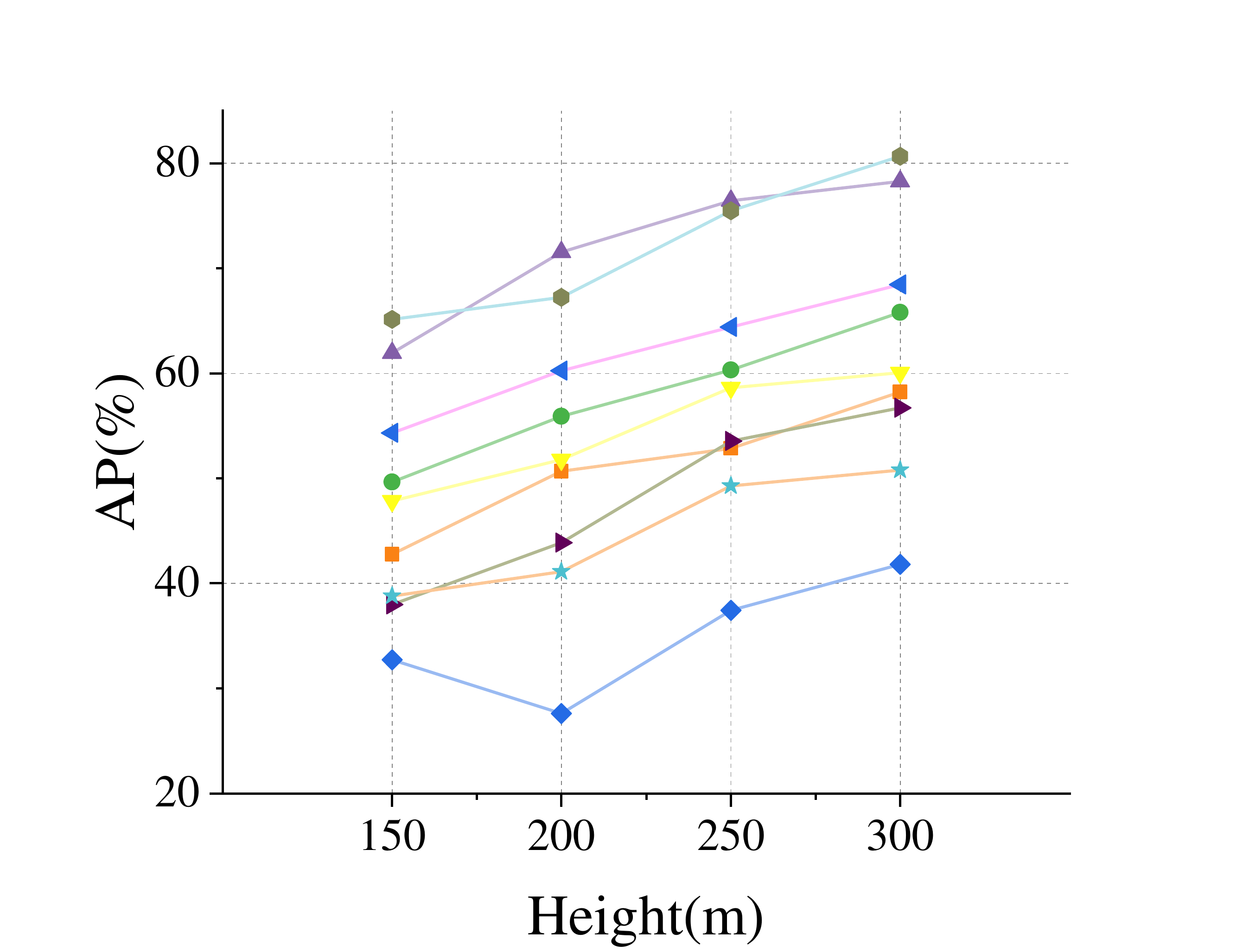}}
\centerline{(c)}
\end{minipage}
\centering
\begin{minipage}[t]{0.24\linewidth}
\centerline{\includegraphics[width=1.3\textwidth]{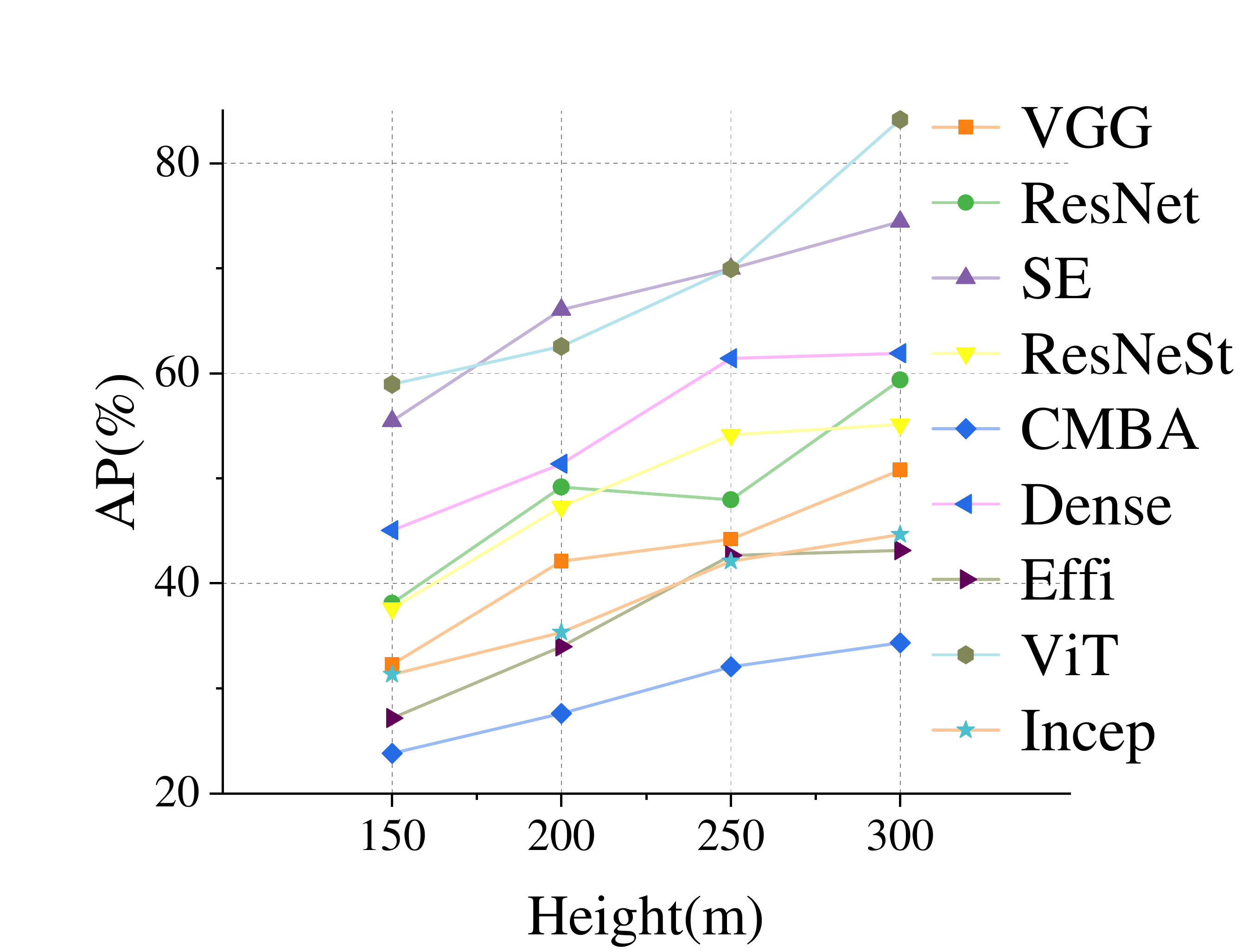}}
\centerline{(d)}
\end{minipage}
\centering
\caption{The Recall@K accuracy curve and AP value curve at 150m, 200m, 250m, and 300m. \ (a): Recall@1 curve of $ \text{Drone}\rightarrow\text{Satellite}$. \ (b):Recall@1 curve of $ \text{Satellite}\rightarrow\text{Drone}$. \ (c):AP curve of $ \text{Drone}\rightarrow\text{Satellite}$. \ (d):AP curve of $ \text{Satellite}\rightarrow\text{Drone}$. \label{acc}}
\end{figure*}

\begin{figure*}[t]
\centering
\begin{minipage}[t]{0.5\linewidth}
\centering
\centerline{\includegraphics[width=1\textwidth]{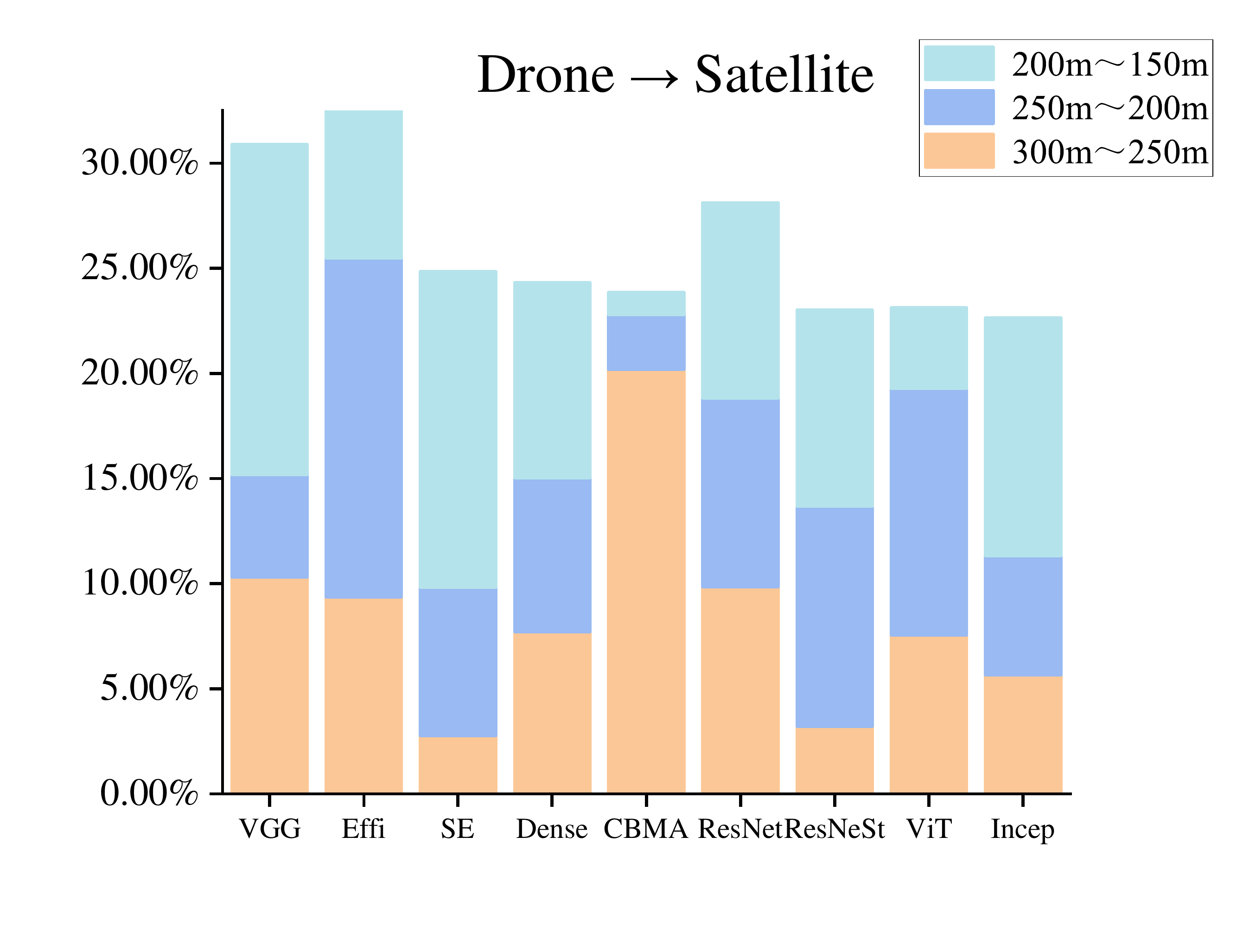}}
\centerline{(a)}
\end{minipage}%
\centering
\begin{minipage}[t]{0.49\linewidth}
\centerline{\includegraphics[width=1\textwidth]{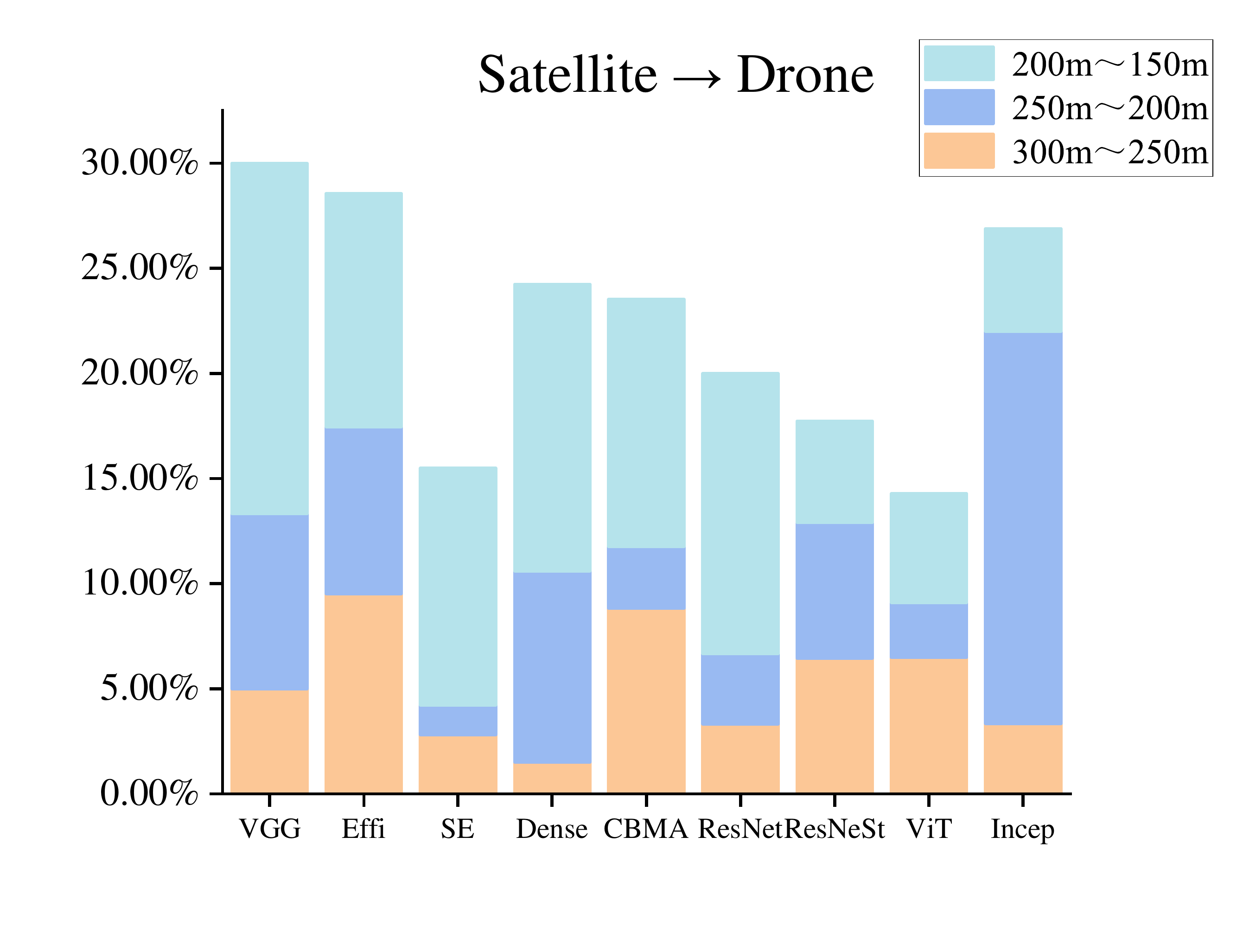}}
\centerline{(b)}
\end{minipage}
\centering
\caption{The robustness of different backbone networks at different Heights. The height of the bars in the bar chart represents the total recall accuracy loss over the height from 300m to 150m. Each of the three colors indicates the loss of accuracy in the respective height interval. \ (a):$ \text{Drone}\rightarrow\text{Satellite}$. \ (b):$ \text{Satellite}\rightarrow\text{Drone}$.} \label{heights}
\end{figure*}

\section{Experiment}
In this \replaced{section}{ chapter},  we first describe the experimental setup and details, followed by a comprehensive evaluation of multiple feature extractors through the pipeline. The impact of multiple queries on the matching performance is explored. In addition, we test the performance of the transfer learning model on SUES-200. Finally, we implement some classical cross-view matching models on SUES-200.

\subsection{Implementation Details}
Different feature extractors are used in our backbone network, and all of them are loaded with ImageNet's pre-trained weights to speed up the convergence of the model. However, the amount of work required to tune so many models to the optimum is considerable. For training, we applied the grid search method to search for the best learning rate, dropout rate, and weight decay hyperparameter values. The image size is resized to (384, 384) before feeding to the network, and only the basic image augmentation methods are used, including random cropping and random horizontal flipping. The optimizer of the neural network is SGD (momentum=0.9), and the initial learning rates of the backbone network and the classification network are set to 0.1 times and 1 time of the learning rate. The learning rate decay is MultiStepLR, and the parameters of the classification network are initialized with Kaiming Initialization \cite{he2015delving}.\added{In the testing stage, we apply \textit{imgaug}\cite{imgaug} to simulate the unfavorable elements for drone view images.} Our model was constructed using the PyTorch framework, and all experiments were conducted on an NVIDIA RTX TiTAN GPU.

\subsection{Evaluation of Different Extractors}
We aim to determine whether SUES-200 can help a model learn highly discriminative features, and whether the pipeline could perform the tasks of training, testing, and evaluation efficiently. In this section, we describe experiments conducted to comprehensively evaluate feature extractors of different DNN architectures and use the model with the best experimental results as the baseline model of SUES-200.

\textbf{Recall and AP.}\  Using the pipeline, we quickly train the models on the SUES-200 for testing and evaluation. As shown in Figure \ref{acc}, we compare the feature extraction capability of different backbone networks by Recall@K and AP. \replaced{In the drone-view target localization task ($\text{Drone}\rightarrow\text{Satellite}$), ViT achieves 59.57\%, 62.30\%, 71.35\%, 77.15\% Recall@1 in four heights, respectively. In the drone navigation task ($\text{Satellite}\rightarrow\text{Drone}$), ViT achieves 82.50\%, 85.00\%, 88.75\%, 96.25\% Recall@1 in four heights, respectively. The performance surpasses the results of other common feature extractors such as ResNet. ViT shows considerable potential for cross-view matching as a general framework. The results also show that as the height of the drone increases, the images captured are less affected by the surrounding environment and the camera field view. The images camera acquired by the camera are more similar to satellite images, and the Recall@K and AP of the model are improved.}{}

 \begin{figure*}[t]
\centering
\begin{minipage}[t]{0.5\linewidth}
\centering
\centerline{\includegraphics[width=1.1\textwidth]{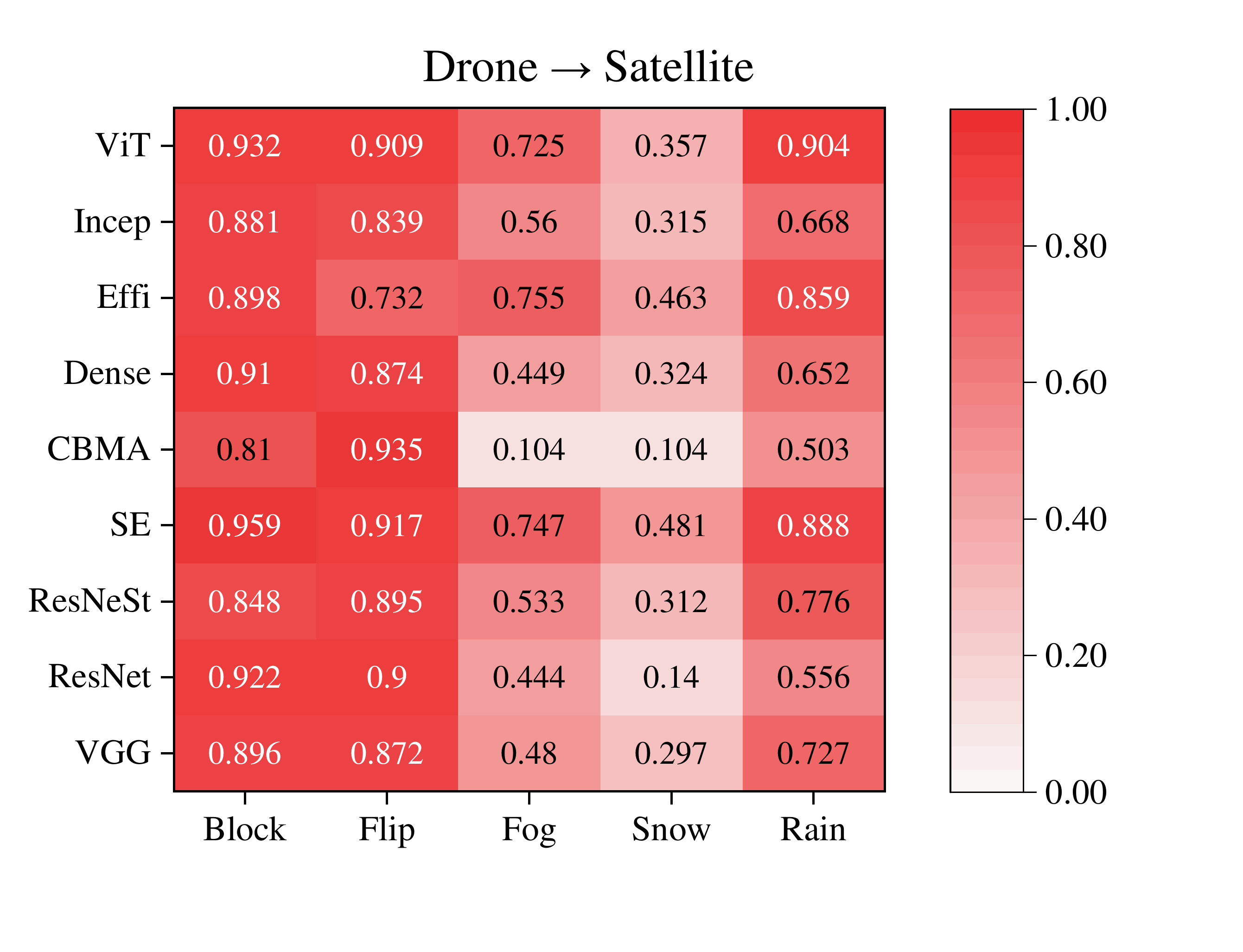}}
\centerline{(a)}
\end{minipage}%
\centering
\begin{minipage}[t]{0.49\linewidth}
\centerline{\includegraphics[width=1.1\textwidth]{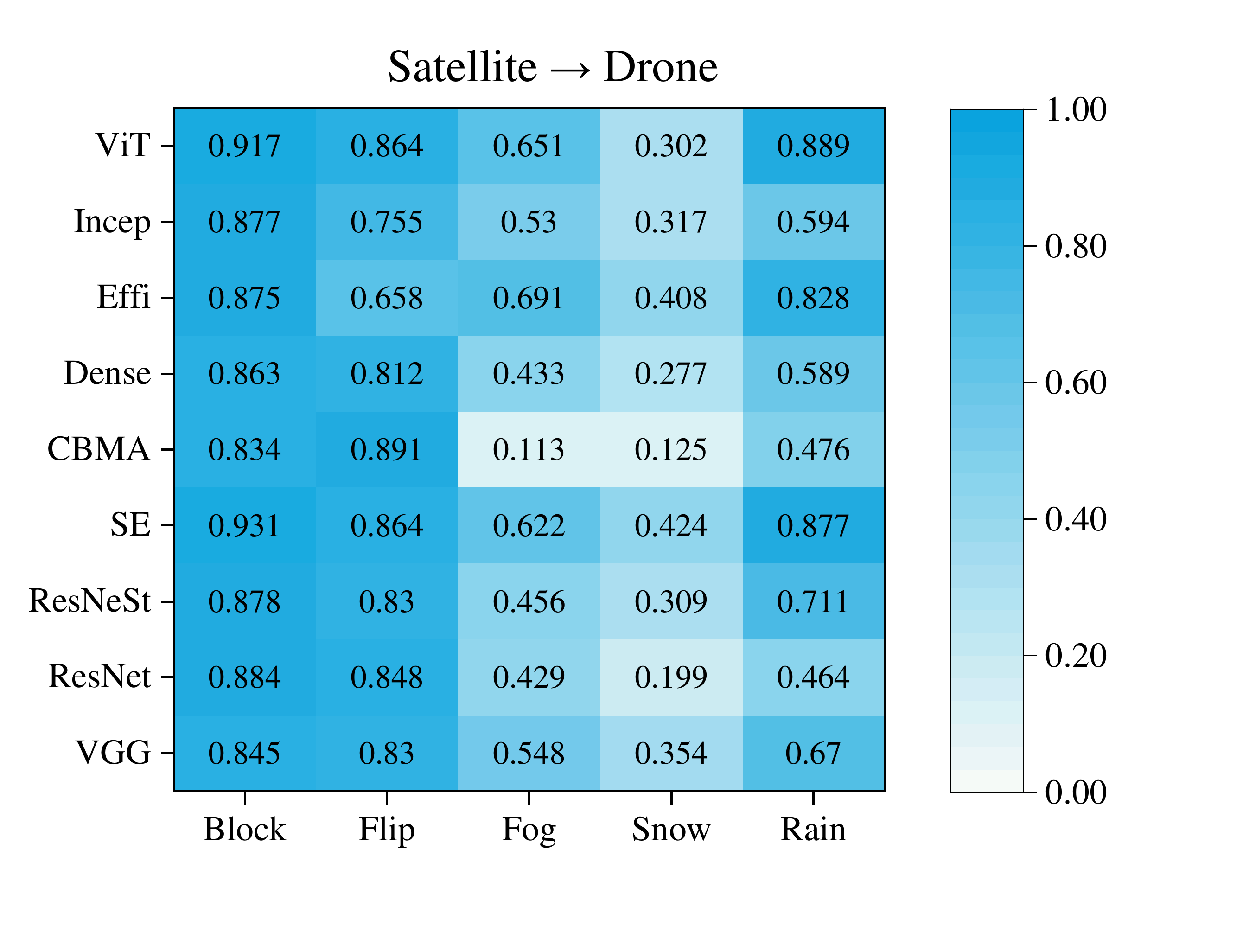}}
\centerline{(b)}
\end{minipage}
\centering
\caption{\added{Heatmaps of the robustness of different backbone networks under unfavorable factors. \ (a):$ \text{Drone}\rightarrow\text{Satellite}$. Darker red indicates that the model was less affected and showed better robustness. \ (b):$ \text{Satellite}\rightarrow\text{Drone}$. Darker blue indicates that the model was less affected and showed better robustness.}} \label{factors}
\end{figure*}

\begin{table*}[t]
\caption{The number of parameters of all models and their inference speeds with benchmark \label{real-time}}
\centering
\begin{tabular}{cccc}
\hline
\multirow{2}{*}{Backbone} & \multirow{2}{*}{Params(M)} & \multirow{2}{*}{$ \text{Drone}\rightarrow\text{Satellite}$} & \multirow{2}{*}{$ \text{Satellite}\rightarrow\text{Drone}$} \\
                          &                            &         &        \\
\hline
\toprule
VGG16-bn                  & 272.86                     & 1.18            & 1.17    \\
ResNet-50                 & 49.24                      & 1.00            & 1.00    \\
SE-ResNet-50               & 54.30                     & 1.02            & 1.02    \\
ResNeSt-50                & 53.09                      & 1.02            & 1.00    \\
CBAM-ResNet-50             & 59.30                     & 1.04            & 1.02    \\
DenseNet-201              & 35.73                      & 1.05            & 1.02    \\
EfficientNetv1-b4         & 37.06                      & 1.01            & 1.00    \\
Inceptionv4               & 83.98                      & 1.03            & 1.01    \\
ViT-base				  & 172.20                     & 2.45            & 2.48    \\
\hline
\end{tabular}
\end{table*}

\textbf{\replaced{Robustness at different heights}{ Robustness}.}\ 
\replaced{Because the height of the drone affects the accuracy of the matching system, we evaluate the robustness of the model to different heights in positioning or navigation tasks using Equations (4)-(5). As may be observed from Figure \ref{heights}, the accuracy of ViT decreased by only 23.13\% and 14.28\% of accuracy on two tasks from 300m to 150m, respectively. This shows strong robustness when the size of the target scene changes. We argue that the self-attention mechanism of Transformer architecture helps ViT ignore the redundant information at low heights.}{}

\added{
\textbf{Robustness to uncertainties.}\ Drones commonly face many negative factors in the outdoor environment. We simulate these situations with image augmentation techniques. Equation (4) is used to evaluate the influence of those factors. Figure \ref{factors} shows the ability of different models to cope with uncertainties. The vertical axis shows that ViT and SE-ResNet exhibit better resistance to uncertainties than other models. The horizontal axis shows that snow was the most difficult factor to overcome. Extracting invariant features from images in wintertime scenes is very challenging; no model is able to reach even the original 50\% AP value under this condition.}

\textbf{\replaced{Inference speed.}{Real-time}}\ In the inference phase, inference speed is a vital evaluation metric, and it also directly determines whether the model can be put into practical application. Therefore, we evaluate the inference speed of different models under two tasks, as may be observed from Table \ref{real-time}. We take the inference time of ResNet as the baseline time: 1.00. \replaced{We find that ViT spent the most time on inference, and Task1 and Task2 were 2.45 and 2.48 times the base time, respectively. Although it has fewer parameters than VGG, the computational complexity of ViT is higher.}{}

\begin{table*}[t]
\caption{The matching accuracy (\%) of multiple queries based on the baseline. 50,25,10,5,1 denote multiple-query image setting\label{multiqueries}}
\centering
\begin{tabular}{lcccccccc}
\hline
\multicolumn{9}{c}{$ \text{Drone}\rightarrow\text{Satellite}$}                                                                                  \\
\hline
\multirow{2}{*}{Query} & \multicolumn{2}{c}{150m} & \multicolumn{2}{c}{200m} & \multicolumn{2}{c}{250m} & \multicolumn{2}{c}{300m} \\
                       & Recall@1     & AP        & Recall@1     & AP        & Recall@1     & AP        & Recall@1     & AP        \\
\hline
\toprule
50                     & 75.00        & 78.99     & 75.00        & 78.50     & 78.75        & 82.27     & 85.00        & 87.41     \\
25                     & 71.25        & 75.81     & 73.13        & 77.00     & 76.25        & 80.10     & 85.63        & 87.58     \\
10                     & 67.75        & 72.68     & 69.00        & 73.50     & 76.25        & 79.95     & 84,25        & 86.57     \\
5                      & 65.37        & 70.49     & 68.25        & 72.72     & 75.50        & 79.22     & 82.00        & 84.75     \\
1                      & 59.32        & 64.93     & 62.30        & 67.24     & 71.35        & 75.49     & 77.17        & 80.67      \\

\hline
\end{tabular}
\end{table*}

\begin{table*}[t]
\caption{Test results of transfer learning models and pre-trained weights on SUES-200. \label{transfer}}
\centering
\begin{tabular}{lcccccccc}
\hline
\multicolumn{9}{c}{$ \text{Drone}\rightarrow\text{Satellite}$}                                                                                                            \\
\hline
\multicolumn{1}{c}{\multirow{2}{*}{Training Set}} & \multicolumn{2}{c}{150m} & \multicolumn{2}{c}{200m} & \multicolumn{2}{c}{250m} & \multicolumn{2}{c}{300m} \\

\multicolumn{1}{c}{}                             & Recall@1     & AP        & Recall@1     & AP        & Recall@1     & AP        & Recall@1     & AP        \\
\hline
\toprule
ImageNet                                         & 13.20         & 17.83     & 16.70        & 22.15     & 13.55        & 17.96     & 14.27        & 18.84     \\
University-1652                                  & 54.90        & 61.11     & 63.55       & 68.82     & 68.53        & 73.20     & 72.00        & 76.29     \\
SUES-200(from scratch)                           & 14.43         & 19.03     & 18.25         & 23.25     & 21.22         & 26.36     & 24.30         & 29.96     \\
SUES-200(ImageNet pre-trained)                    & 59.32        & 64.93     & 62.30        & 67.24     & 71.35        & 75.49     & 77.17       & 80.67     \\
SUES-200(U1652 pre-trained)                       & 71.67        & 75.55     & 75.57        & 78.97     & 79.97        & 82.50     & 81.42        & 84.11     \\
\hline
\multicolumn{9}{c}{$ \text{Satellite}\rightarrow\text{Drone}$}                                                                                                            \\
\hline
\multicolumn{1}{c}{\multirow{2}{*}{Training Set}} & \multicolumn{2}{c}{150m} & \multicolumn{2}{c}{200m} & \multicolumn{2}{c}{250m} & \multicolumn{2}{c}{300m} \\
\multicolumn{1}{c}{}                             & Recall@1     & AP        & Recall@1     & AP        & Recall@1     & AP        & Recall@1     & AP        \\
\hline
\toprule
ImageNet                                         & 16.25        & 9.85      & 7.50         & 6.38      & 18.75        & 11.96     & 26.25        & 16.00     \\
University-1652                                  & 61.25        & 48.08     & 75.00        & 60.24     & 77.50        & 66.51     & 75.00        & 70.29    \\
SUES-200(from scratch)                           & 17.50        & 11.62      & 30.00        & 18.56     & 35.00       & 22.13    & 47.50      & 29.46      \\
SUES-200(ImageNet pre-trained)                    & 82.50        & 58.95     & 85.00        & 62.56     & 88.75        & 69.96     & 96.25        & 84.16      \\
SUES-200(U1652 pre-trained)                       & 85.00        & 71.36     & 86.25        & 75.96     & 88.75        & 79.54     & 92.50        & 84.89    \\
\hline
\end{tabular}
\end{table*}

\subsection{Multiple Queries}
\textbf{We consider whether multi-angle feature fusion improves the efficiency of matching.}\ In previous matching experiments, a single drone-view image was used as a query for $ \text{Drone}\rightarrow\text{Satellite}$. Each scene in the SUES-200 dataset provides a full 360-degree view of the drone view image, which provides complete and comprehensive information about the target scene. Therefore, if a single query cannot describe the target scene, we can introduce multiple angles of drone view images as queries. as may be observed from Table \ref{multiqueries}, we set the multiple-query image to {50, 25, 10, 5, 1}. The experimental results show that the multiple queries contain more images, and the Recall@K and AP of the matching are enhanced accordingly. When the average features of 50 images are used as queries, \replaced{the accuracy of Recall@1 is generally improved by 15\%, compared with the single query. }{}

\subsection{Transfer Learning}
\textbf{We consider whether previous datasets can help a model learn features at different heights, and whether pre-trained weights exhibit an impact on training process.}\  We test whether the models obtained from training on the ImageNet dataset, as well as the University-1652 dataset, can extract discriminative features at different heights. We train a model from scratch on SUES-200 with ImageNet and University-1652 as pre-trained weights. The backbone networks in the above models are ViT. As shown in Table \ref{transfer}, the University-1652-based transfer learning model achieves surprising results compared to ImageNet, which validates that University-1652 can be applied to real scenes. But University-1652's ability at different heights is still limited because the dataset does not distinguish the effects of different heights. Further, we find that the model trained from scratch is much less capable of extracting features than the model trained based on ImageNet. Another interesting finding is that the initial training process of the model based on the pre-trained weights of University-1652 performed better than the one based on ImageNet, which also shows that the initialization weights of the model are significantly important.

\subsection{Other Baseline Models on SUES-200}
\textbf{We evaluate the performance of the classical cross-view matching model on the SUES-200 dataset.}\  Some previous works\cite{ding2021practical,wang2021each} have designed deep neural networks that achieved excellent performance on different cross-view matching datasets. We select LCM\cite{ding2021practical} and LPN\cite{wang2021each} and migrate their backbone network designs into our pipeline for training. The experimental results are shown in Table \ref{others}. Due to the feature partitioning strategy presented by the LPN for extracting semantic information, the strategy is able to extract global features of the image instead of focusing on the center of the image alone. LPN achieves competitive performance on SUES-200, especially in Task 1.  

\begin{table*}[t]
\caption{Test performances of LCM and LPN on SUES-200 \label{others}}
\centering
\begin{tabular}{lcccccccccc}
\hline
\multicolumn{9}{c}{$ \text{Drone}\rightarrow\text{Satellite}$}                                                                                                        \\
\hline
\multicolumn{1}{c}{\multirow{2}{*}{Methods}} & \multicolumn{2}{c}{150m} & \multicolumn{2}{c}{200m} & \multicolumn{2}{c}{250m} & \multicolumn{2}{c}{300m} \\
\multicolumn{1}{c}{}                         & Recall@1     & AP        & Recall@1     & AP        & Recall@1     & AP        & Recall@1     & AP        \\
\hline
\toprule
SUES-200   baseline                          & 59.32        & 64.93     & 62.30        & 67.24     & 71.35        & 75.49     & 77.17       & 80.67     \\
LCM   \cite{ding2021practical}               & 43.42        & 49.65     & 49.42        & 55.91     & 54.47        & 60.31     & 60.43        & 65.78     \\
LPN(block=4) \cite{wang2021each}             & 61.58        & 67.23     & 70.85        & 75.96     & 80.38        & 83.80     & 81.47        & 84.53     \\
\hline
\multicolumn{9}{c}{$ \text{Satellite}\rightarrow\text{Drone}$}                                                                                                        \\
\hline
\multicolumn{1}{c}{\multirow{2}{*}{Methods}} & \multicolumn{2}{c}{150m} & \multicolumn{2}{c}{200m} & \multicolumn{2}{c}{250m} & \multicolumn{2}{c}{300m} \\

\multicolumn{1}{c}{}                         & Recall@1     & AP        & Recall@1     & AP        & Recall@1     & AP        & Recall@1     & AP        \\
\hline
\toprule
SUES-200   baseline                          & 82.50        & 58.95     & 85.00        & 62.56     & 88.75        & 69.96     & 96.25        & 84.16      \\
LCM\cite{ding2021practical}                  & 57.50        & 38.11     & 68.75        & 49.19     & 72.50        & 47.94     & 75.00        & 59.36     \\
LPN(block=4)\cite{wang2021each}              & 83.75        & 66.78     & 88.75        & 75.01     & 92.50        & 81.34     & 92.50        & 85.72    \\
\hline
\end{tabular}
\end{table*}

\begin{table*}[t]
\caption{\added{Ablation study of different Feature dimensions on the SUES-200 dataset.}\label{size}}
\centering
\begin{tabular}{lccccccccc}
\hline
\multicolumn{10}{c}{$ \text{Drone}\rightarrow\text{Satellite}$} \\
\hline
\multicolumn{1}{c}{\multirow{2}{*}{Feature Dimension}} & \multicolumn{2}{c}{150m} & \multicolumn{2}{c}{200m} & \multicolumn{2}{c}{250m} & \multicolumn{2}{c}{300m}  & \multirow{2}{*}{Height Robustness} \\
\multicolumn{1}{c}{}                            & Recall@1     & AP        & Recall@1     & AP        & Recall@1     & AP        & Recall@1     & AP           &                             \\
\hline
\toprule
256                                             & 48.15        & 54.95     & 57.80        & 63.14     & 65.13        & 70.26     & 70.15       & 75.03         & 31.36\%                      \\
512                                             & 59.32        & 64.93     & 62.30        & 67.24     & 71.35        & 75.49     & 77.17        & 80.67         & 23.13\%                      \\
1024                                            & 51.13        & 56.71     & 64.35        & 69.09     & 72.25        & 76.34     & 76.63        & 80.24         & 33.28\%                       \\
\hline
\multicolumn{10}{c}{$ \text{Satellite}\rightarrow\text{Drone}$} \\
\hline
\multicolumn{1}{c}{\multirow{2}{*}{Feature Dimension}} & \multicolumn{2}{c}{150m} & \multicolumn{2}{c}{200m} & \multicolumn{2}{c}{250m} & \multicolumn{2}{c}{300m}  & \multirow{2}{*}{Height Robustness} \\
\multicolumn{1}{c}{}                            & Recall@1     & AP        & Recall@1     & AP        & Recall@1     & AP        & Recall@1     & AP                 &                             \\
\hline
\toprule
256                                             & 63.75        & 48.86     & 77.50        & 60.81     & 83.75        & 69.23     & 86.25        & 71.54               & 26.09\%                      \\
512                                             & 82.50        & 58.95     & 85.00        & 62.56     & 88.75        & 69.96     & 96.25        & 84.16               & 14.29\%                     \\
1024                                            & 72.50        & 52.75     & 83.75        & 67.52     & 88.75        & 76.77     & 90.00        & 76.46               & 19.44\%                      \\
\hline
\end{tabular}
\end{table*}

\section{Ablation study}

\subsection{\replaced{Effect of Feature Dimensions}{}}
\replaced{
\textbf{We also consider how different feature dimensions affect the model.}\  The dimensionality of features extracted from the drone and satellite images in the SUES-200 dataset was 512, as shown in Figure \ref{network}, FC1. Therefore, in the ablation learning phase, we reset the feature dimension to 256 and 1024, keeping all other conditions constant. As shown in Table \ref{size}, when we set the dimension to 256, the Recall@1 and AP accuracy both decreased. When we set the dimension to 1024, the performance is better than with 512 dimensions in some metrics. However, 512 is still the overall optimal size and we thus use this dimensionality in the baseline model.}{}

\begin{figure*}[t]
\centering
\begin{minipage}[t]{0.45\linewidth}
\centering
\centerline{\includegraphics[width=1.0\textwidth]{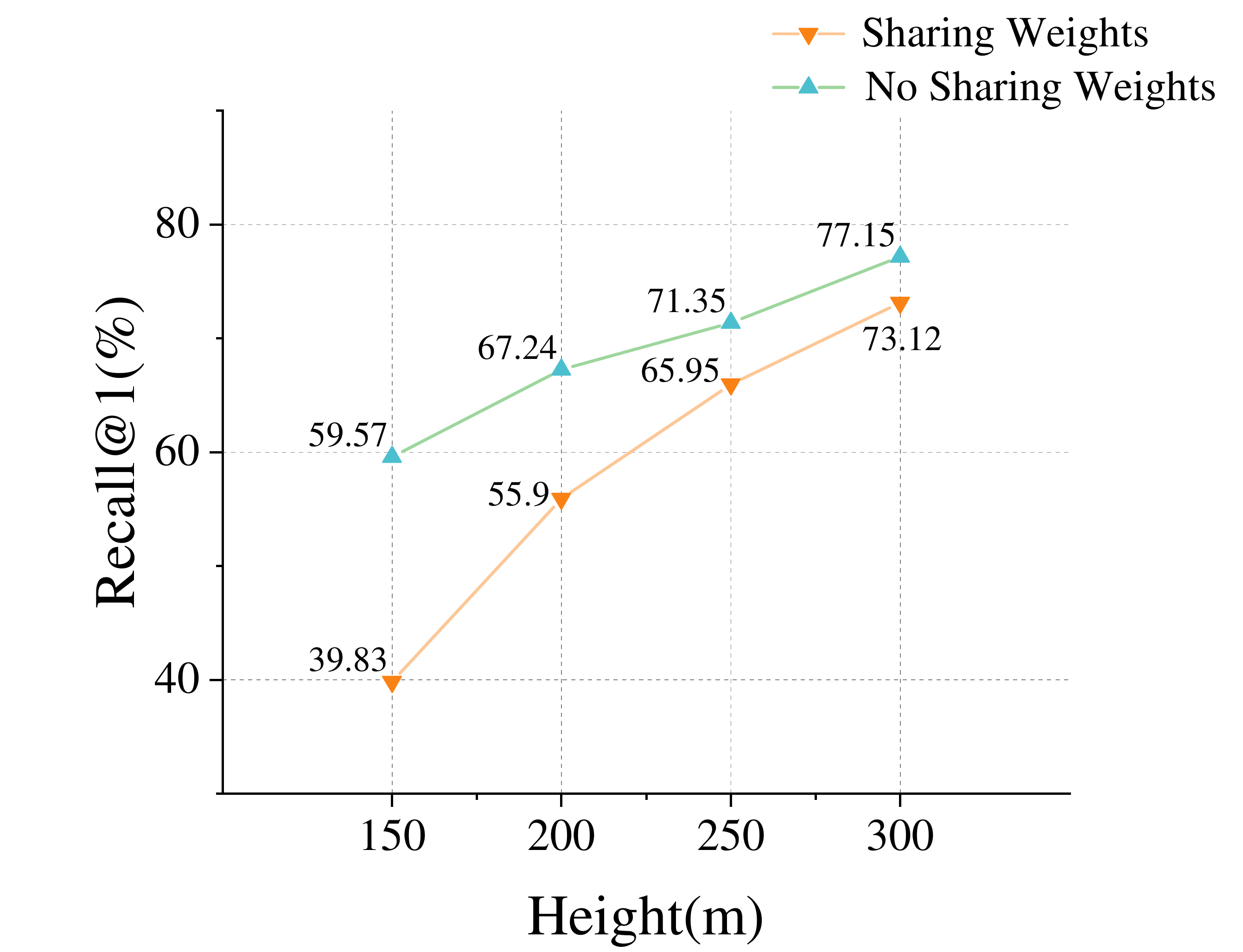}}
\centerline{(a)}
\end{minipage}%
\centering
\begin{minipage}[t]{0.45\linewidth}
\centerline{\includegraphics[width=1.0\textwidth]{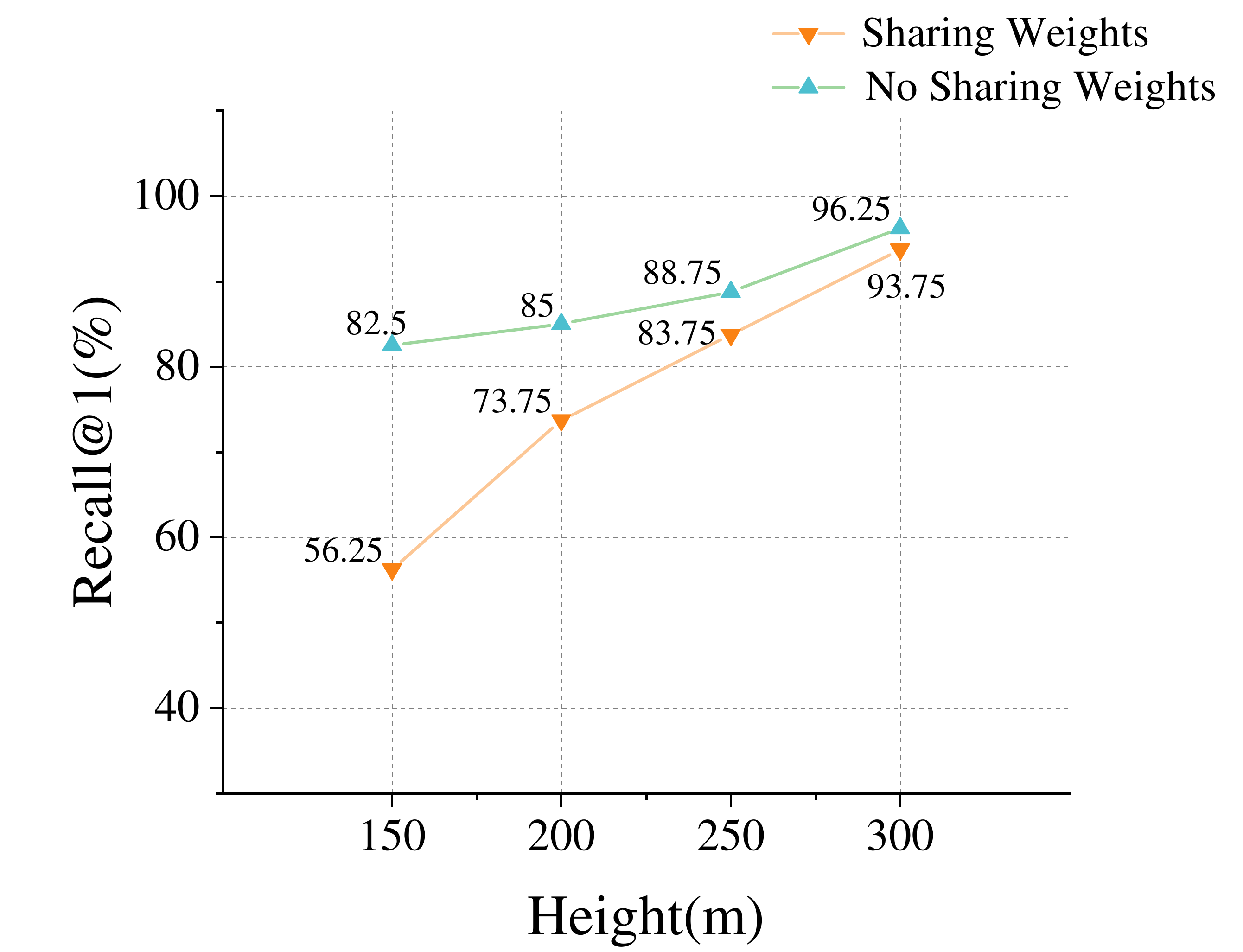}}
\centerline{(b)}
\end{minipage}
\centering
\caption{The accuracy of Recall@1 without sharing weights is always higher than that of Recall@1 with sharing weights, but the gap decreases as the height rises. (a) $\text{Drone}\rightarrow\text{Satellite}$ (b)$\text{Satellite}\rightarrow\text{Drone}$ \label{share}}
\end{figure*}

\begin{table*}[t]
\caption{Ablation study of different loss terms\label{loss}}
\centering
\begin{tabular}{lccccccccc}
\hline
\multicolumn{10}{c}{$ \text{Drone}\rightarrow\text{Satellite}$}                                                                                                                                        \\
\hline
\multicolumn{1}{c}{\multirow{2}{*}{Loss}}       & \multicolumn{2}{c}{150m} & \multicolumn{2}{c}{200m} & \multicolumn{2}{c}{250m} & \multicolumn{2}{c}{300m} & \multirow{2}{*}{Height Robustness} \\
\multicolumn{1}{c}{}                            & Recall@1     & AP        & Recall@1     & AP        & Recall@1     & AP        & Recall@1     & AP        &                             \\
\hline
\toprule
CrossEntropy  \cite{hadsell2006dimensionality}  & 59.32        & 64.93     & 62.30        & 67.24     & 71.35        & 75.49     & 77.17        & 80.67     & 23.13\%                        \\
Contrastive      \cite{hoffer2015deep}          & 54.03        & 59.85     & 59.82        & 63.12     & 67.28        & 71.58     & 71.60        & 75.79     & 24.53\%                      \\
Triplet(margin=0.3)                             & 51.34        & 57.36     & 63.57        & 68.49     & 68.62        & 73.13     & 71.72        & 75.69     & 28.41\%                       \\
\hline
\multicolumn{10}{c}{$ \text{Satellite}\rightarrow\text{Drone}$}                                                                                                                                        \\
\hline
\multicolumn{1}{c}{\multirow{2}{*}{Loss}} & \multicolumn{2}{c}{150m} & \multicolumn{2}{c}{200m} & \multicolumn{2}{c}{250m} & \multicolumn{2}{c}{300m} & \multirow{2}{*}{Robustness} \\
\multicolumn{1}{c}{}                            & Recall@1     & AP        & Recall@1     & AP        & Recall@1     & AP        & Recall@1     & AP        &                             \\
\hline
\toprule
CrossEntropy \cite{hadsell2006dimensionality}   & 82.50        & 58.95     & 85.00        & 62.56     & 88.75        & 69.96     & 96.25        & 84.16    & 14.29\%                      \\
Contrastive   \cite{hoffer2015deep}             & 75.00 	       & 58.06     & 81.25        & 60.31     & 86.25        & 70.97     & 91.25        & 73.14     & 17.81\%                       \\
Triplet(margin=0.3)                             & 75.00        & 56.95     & 81.25        & 59.52     & 85.00        & 68.09     & 87.50        & 77.08     & 14.29\%                  \\
\hline
\end{tabular}
\end{table*}

\subsection{Effects of sharing weights}
\textbf{We consider whether sharing weights could help the model learn better features.} With increasing height, drone and satellite images become more and more similar. Hence, the question arises as to whether the model learning efficiency of the model can be improved by sharing the weights of the backbone. We test the effects of sharing model weights on the final test results in the baseline model. Figure \ref{share} shows that the evaluation metrics of both tasks show significant decreases when the sharing weights are available. Still, the difference values between the shared and unshared weights decreased with height. Images collected with increasing height are more similar to satellite images. One possible explanation is that sharing weights can help the model extract more efficient features in similar image pairs.

\subsection{Effects of different loss function}
\textbf{We also consider whether other loss functions would affect the learning performance of the model.}\ The most common loss functions in previous studies of matching retrieval tasks are contrastive loss\cite{hadsell2006dimensionality} and triplet loss\cite{hoffer2015deep}, and these loss functions achieve good performance in other works, such as ReID. To verify the feasibility of these loss functions on our baseline, we strictly hold the backbone network and other parameters constant during the experiments. From Table \ref{loss}, it may be observed that each of these three loss functions shows advantages and disadvantages in terms of Recall@K and AP. \replaced{However, when evaluating robustness to height, the accuracy of cross-entropy loss fall the least accuracy from 300m to 150m.}{}

\begin{table*}[t]
\caption{\added{Ablation study of distance measurement algorithms on SUES-200}\label{distance}}
\centering
\begin{tabular}{lcccccccc}
\hline
\multicolumn{9}{c}{$ \text{Drone}\rightarrow\text{Satellite}$}                                                                    \\
\hline
\multicolumn{1}{c}{\multirow{2}{*}{Distance}}       & \multicolumn{2}{c}{150m} & \multicolumn{2}{c}{200m} & \multicolumn{2}{c}{250m} & \multicolumn{2}{c}{300m}  \\
\multicolumn{1}{c}{}        & Recall@1     & AP        & Recall@1     & AP        & Recall@1     & AP        & Recall@1     & AP        \\
\hline
\toprule
Euclidean    				& 59.70        & 65.24     & 62.17        & 67.13     & 71.30        & 75.46     & 77.28       & 80.75                           \\
Manhattan   				& 57.30        & 62.98     & 61.83        & 66.78     & 69.10        & 73.61     & 75.52       & 79.38                          \\
Cosine      			    & 59.32        & 64.93     & 62.30        & 67.24     & 71.35        & 75.49     & 77.17        & 80.67                          \\
\hline
\multicolumn{9}{c}{$ \text{Satellite}\rightarrow\text{Drone}$}                                                                                      \\
\hline
\multicolumn{1}{c}{\multirow{2}{*}{Distance}} & \multicolumn{2}{c}{150m} & \multicolumn{2}{c}{200m} & \multicolumn{2}{c}{250m} & \multicolumn{2}{c}{300m}   \\
\multicolumn{1}{c}{}        & Recall@1     & AP        & Recall@1     & AP        & Recall@1     & AP        & Recall@1     & AP       \\
\hline
\toprule
Euclidean    & 81.25        & 58.93     & 85.00        & 62.55     & 90.00        & 69.98     & 95.00        & 84.14                           \\
Manhattan    & 76.25 	    & 56.80     & 85.00        & 59.69     & 88.75        & 68.29     & 92.50        & 81.98                           \\
Cosine       & 82.50        & 58.95     & 85.00        & 62.56     & 88.75        & 69.96     & 96.25        & 84.16                         \\
\hline
\end{tabular}
\end{table*}

\subsection{\added{Effects of Distance Measurement Algorithm}}
\added{
\textbf{We consider whether different distance measurement algorithms affect the matching results.}\ In cross-view matching, several measurement algorithms exist, such as euclidean distance, manhattan distance, and cosine distance. We apply cosine distance in our baseline model due to its good performance in image retrieval tasks\cite{zhu2020voc, luo2019bag}. How do other distance measurement algorithms perform on SUES-200? As shown in Table \ref{distance}, Manhattan distance has the worst performance compared with other distance measurement algorithms. Euclidean distance achieves comparable results to the cosine distance.}

\subsection{\added{Effects of Distractors in Gallery}}
\added{
\textbf{We also consider whether distractors affect the matching process.}\ In the test stage, we add training data to the gallery set as distractors. To compare the performance of the model without distractors, we remove training data from the gallery set. It can be seen in Table \ref{distractors}, the model's performance improves significantly under the gallery set without distractors. The absence of distractors made it easier for the model to find the correct match. Therefore, we believe a gallery set with more data allows for a more comprehensive evaluation of the model.}

\begin{table*}[t]
\caption{\added{Ablation study of distractors on SUES-200}\label{distractors}}
\centering
\begin{tabular}{lcccccccc}
\hline
\multicolumn{9}{c}{$ \text{Drone}\rightarrow\text{Satellite}$}                                                                    \\
\hline
\multicolumn{1}{c}{\multirow{2}{*}{Gallery set}}       & \multicolumn{2}{c}{150m} & \multicolumn{2}{c}{200m} & \multicolumn{2}{c}{250m} & \multicolumn{2}{c}{300m}  \\
\multicolumn{1}{c}{}        & Recall@1     & AP        & Recall@1     & AP        & Recall@1     & AP        & Recall@1     & AP        \\
\hline
\toprule
No Distractors    			& 72.03        & 76.12     & 73.18        & 77.10     & 80.48        & 83.31     & 83.17       & 85.90                           \\
Distractors   				& 59.32        & 64.93     & 62.30        & 67.24     & 71.35        & 75.49     & 77.17        & 80.67                          \\
\hline
\multicolumn{9}{c}{$ \text{Satellite}\rightarrow\text{Drone}$}                                                                                      \\
\hline
\multicolumn{1}{c}{\multirow{2}{*}{Gallery set}} & \multicolumn{2}{c}{150m} & \multicolumn{2}{c}{200m} & \multicolumn{2}{c}{250m} & \multicolumn{2}{c}{300m}   \\
\multicolumn{1}{c}{}        & Recall@1     & AP        & Recall@1     & AP        & Recall@1     & AP        & Recall@1     & AP       \\
\hline
\toprule
No Distractors    			& 88.75        & 71.65     & 90.00        & 70.93     & 91.25        & 76.83     & 95.00        & 88.29                           \\
Distractors   				& 82.50        & 58.95     & 85.00        & 62.56     & 88.75        & 69.96     & 96.25        & 84.16                          \\
\hline
\end{tabular}
\end{table*}

\begin{table*}[t]
\caption{\added{Effect of adding the losses}\label{losses}}
\centering
\begin{tabular}{lcccc}
\hline
\multicolumn{5}{c}{$ \text{Drone}\rightarrow\text{Satellite}$}                                                 \\
\multicolumn{1}{c}{FC layer weight}     & Recall@1     & Recall@5        & Recall@10     & AP               			\\
\hline
\toprule
Adding losses jointly optimize  				& 56.01        & 80.12     & 91.18        & 62.20                               \\
Separated losses independently optimize  		 & 4.65        & 12.00     & 20.65        & 7.62                              \\
Satellite init									& 52.97        & 78.72     & 89.10        & 58.87     							\\
Drone init										& 49.20        & 76.40     & 87.55        & 55.42    					 		\\
\hline
\multicolumn{5}{c}{$ \text{Satellite}\rightarrow\text{Drone}$}                                               	\\
\multicolumn{1}{c}{FC layer weight}     & Recall@1     & Recall@5        & Recall@10     & AP               			\\
\hline
\toprule
Adding losses jointly optimize     						& 75.00        & 86.12     & 89.58        & 55.51                               \\
Separated losses independently optimize    				& 5.00         & 7.50      & 8.75         & 6.17                              \\
Satellite init											& 62.50        & 70.00     & 71.25        & 53.87     							\\
Drone init												& 50.00        & 57.50     & 63.50        & 44.57   					 		\\
\hline
\end{tabular}
\end{table*}

\begin{table*}[!t]
\caption{\added{Effect of different ensemble strategies}\label{ensemble}}
\centering
\begin{tabular}{lcccccccc}
\hline
\multicolumn{9}{c}{$ \text{Drone}\rightarrow\text{Satellite}$}                                                                                  \\
\hline
\multirow{2}{*}{Method} & \multicolumn{2}{c}{150m} & \multicolumn{2}{c}{200m} & \multicolumn{2}{c}{250m} & \multicolumn{2}{c}{300m} \\
                       & Recall@1     & AP        & Recall@1     & AP        & Recall@1     & AP        & Recall@1     & AP        \\
\hline
\toprule
Average                    & 66.25        & 71.61     & 77.50        & 80.84     & 80.00        & 83.84     & 82.50        & 85.49     \\    
Max pooling                & 58.74        & 65.19     & 72.50        & 76.19     & 66.25        & 72.37     & 77.50         & 80.92     \\
Voting                     & 55.00        & 62.46     & 71.25        & 75.15     & 76.25        & 79.95     & 84,25        & 86.57     \\
\hline
\end{tabular}
\end{table*}

\begin{figure}[t]
  \begin{center}
  \includegraphics[width=0.4\textwidth]{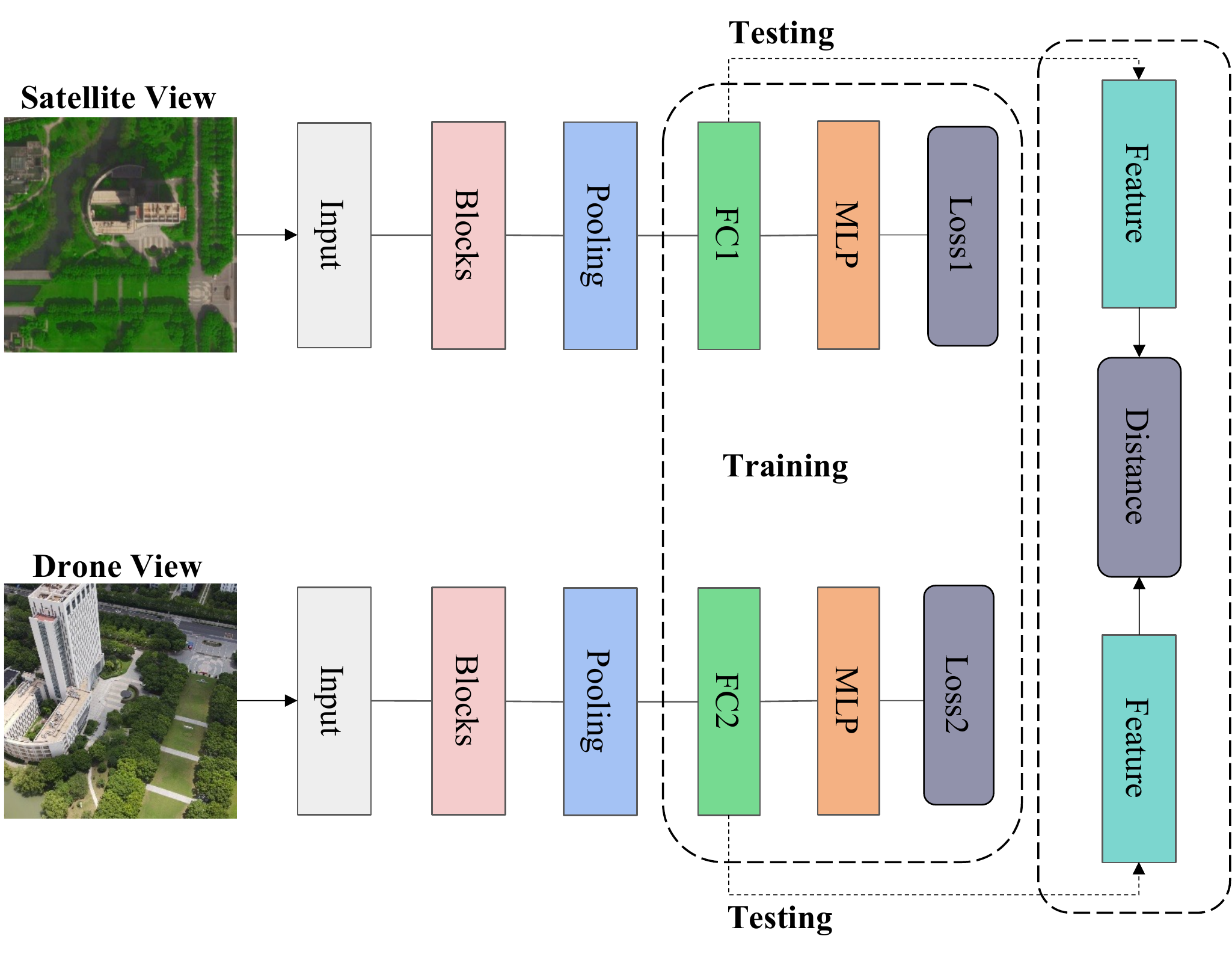}\\
  \caption{The structure of the model for calculating the loss separately}\label{single}
  \end{center}
\end{figure}

\subsection{\added{Effects of adding the losses}}
\added{
\textbf{We consider whether adding the losses would affect model.}
To verify the effect of adding the losses from two branches, we divide the network into two parts in the training stage. As shown in Figure \ref{single}, we train two SE-ResNet50 models at 150m height. Their loss functions are no longer added together, and their models were optimized independently. Experimental results are shown in Table \ref{losses}; the performance of both tasks considerably declined. Why do the results exceed our expectations? How is the current network model different from the previous one? The key is the first FC layer (See FC1 in Figure \ref{network}), which shares weights with two branches in the original network. If we also share FC layer weights when testing the current network, the experimental results are presented in Table \ref{losses}. we initialize FC1 and FC2 with the weight of either view in the current network (Figure \ref{single}), the result is another huge improvement compared to the former one. }

\added{
Finally, we summarize the potential advantages of adding losses. Firstly, we believe the two branches constrain each other to jointly optimize the FC Layer, resulting in the FC layer being able to extract features available to both views. Secondly, a complete network system is more suitable for adjusting the hyperparameters in the training stage.}

\subsection{\added{Effects of different ensemble strategies}}
\added{
\textbf{We also consider whether different ensemble strategies would affect the multiple queries.} Apart from the numerical average, we also try other ensemble strategies. We employ max pooling and voting to fuse features. Max pooling is a common fusion method. We present it to extract the most remarkable part of the feature maps. The voting method is to select the one with the most occurrences among the predicted labels as the prediction. We apply SE-ResNet-50 as the backbone to evaluate those strategies. The number of query images is 50. It can be seen in Table \ref{ensemble}, we observe that the numerical average arrives better performance than Max pooling and voting on both tasks. We also notice that as the height rises and more predictions become correct, the performance of the voting method also rises.}

\section{Visualization}
\begin{figure*}[t]
  \begin{center}
  \includegraphics[width=1.0\textwidth]{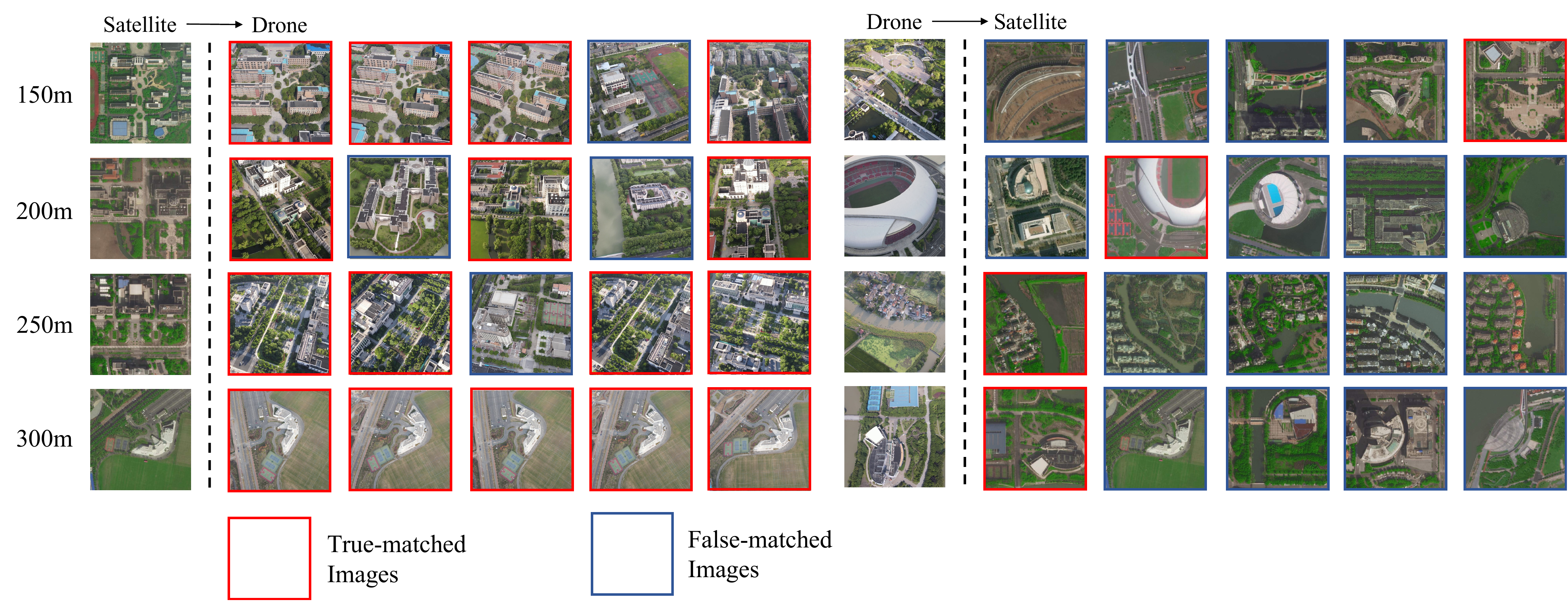}\\
  \caption{Qualitative image retrieval results. Top-5 retrieval results of drone view target localization on SUES-200. Top-5 retrieval results of drone navigation on SUES-200.}\label{visual}
  \end{center}
\end{figure*}

\begin{figure}[t]
  \begin{center}
  \includegraphics[width=0.49\textwidth]{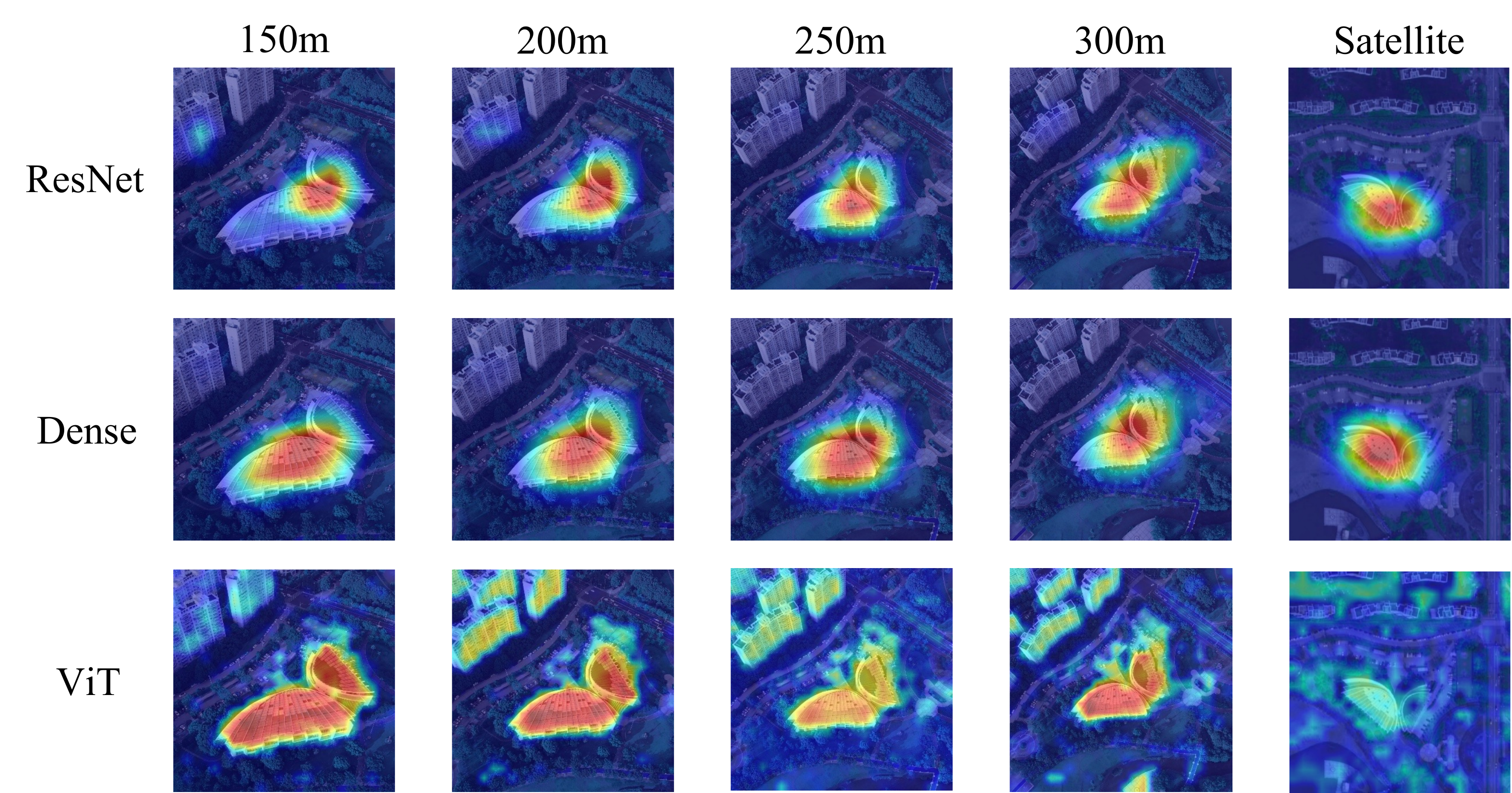}\\
  \caption{Visualization of heatmaps. Heatmaps are generated by ResNet, Dense and Baseline based model}\label{heat}
  \end{center}
\end{figure}

Figure \ref{visual} shows the visualization results of the baseline model under Rank 5 at different heights and two tasks. \replaced{It may be observed that ViT (baseline) was able to retrieve the correct result in some very similar scenes.}{} Furthermore, we also visualize the heat maps generated by different models on SUES-200. Figure \ref{heat} compares the results of ResNet, Dense, and \replaced{ViT}{} on the drone view and satellite view. \replaced{We observe that CNN-based models focused attention on the main target. ViT was also able to notice other things in the background around the target scene. It even had the capability to accurately depict the shape of the target.}{}

\section{Discussion}
In this study, we find that height exhibited a significant effect on cross-view matching. At the heights of 150m and 200m, the drone footage was more influenced by the surrounding environment and the camera pose. The size of the target scene size leads to drone images being very different from the satellite view images. Hence, accuracy at low heights is relatively low. However, with increasing height, the drone is less influenced by redundant information. The accuracy of matching gradually increases. At the same time, we consider that the bottleneck of previous research on cross-view matching studies lies in the lack of a suitable feature extractor. As shown in Table \ref{survey}, most methods are based on the same feature extractor, ResNet. 

\replaced{To test the performance of these feature extractors in a complete way, we also evaluate their performance in the other three aspects through the pipeline. The results show that the ViT-based model has better robustness at different heights and uncertainties. However, it is very challenging for most approaches to extract invariant features in wintertime. Furthermore, the inference speed of the ViT-based model still needs to be improved. Another limitation is that SUES-200 still suffers from a small number of samples and limited viewpoints of the same location. }{}

\section{Conclusion}
In this study, we have investigated the problem of image matching across drone and satellite views at different heights. We have proposed a multi-height, multi-scene benchmark called SUES-200, which contains images collected from aerial drones and satellite images for 200 locations. We have also presented three metrics with a pipeline to comprehensively evaluate the model's ability. 1) robustness at different heights; 2) robustness to uncertainties; 3) inference speed; The results of our experiments have shown that the accuracy and precision of matching increase as the drone's height increases. After evaluating different feature extractors, we provide the model with the best overall performance as the baseline model of SUES-200. For $ \text{Drone}\rightarrow\text{Satellite}$, baseline achieves 59.32, 62.30, 71.35, and 77.17 Recall@1 accuracy at heights of 150m, 200m, 250m, and 300m, respectively. For $ \text{Satellite}\rightarrow\text{Drone}$, baseline achieves 82.50, 85.00, 88.75, and 96.25 Recall@1 accuracy in 150m, 200m, 250m, and 300m, respectively. We also observe that appropriate pre-trained weights and multiple queries can benefit the model to achieve even better performance, which provides an approach to improve matching efficiency further. 

In the future, the main issue to be considered is how to filter out the invalid redundant information at low heights. We also plan to develop a network to adapt to different flight conditions, when the cross-view matching system faces uncertainties. Moreover, the development of a lightweight Transformer architecture for cross-view matching would also be very beneficial to the application of this technology. The data of SUES-200 will also be extended in future research.

\bibliographystyle{IEEEtran}
\bibliography{ref}

\begin{IEEEbiography}
[{\includegraphics[width=1in,height=1.25in,clip,keepaspectratio]{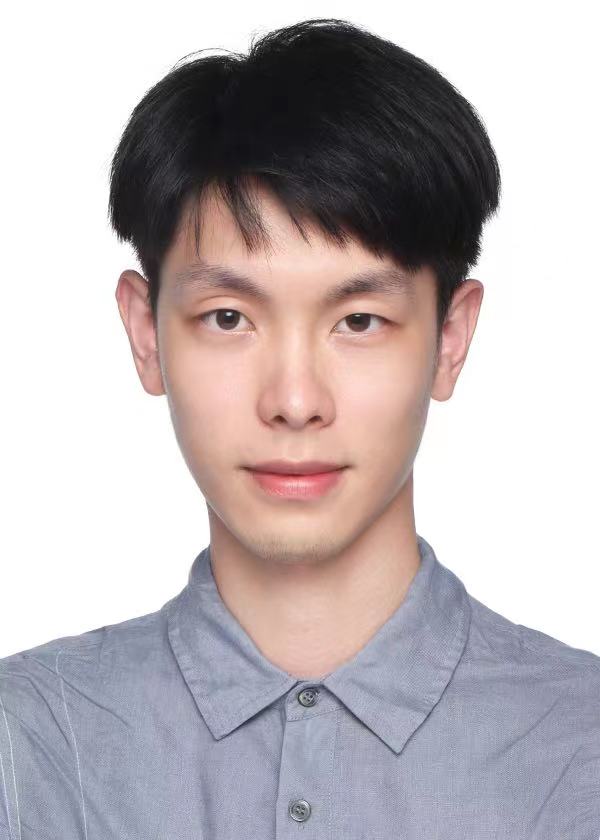}}]{Runzhe Zhu}
received the B.S. degree in Zhejiang Shuren University from Zhejiang, Hangzhou, China, in 2020. He is currently an M.S.student with the department of electrical and electronic engineering of Shanghai University of Engineering Science, Shanghai, China. His research interests include visual geo-localization and cross-view matching.
\end{IEEEbiography}

\begin{IEEEbiography}
[{\includegraphics[width=1in,height=1.25in,clip,keepaspectratio]{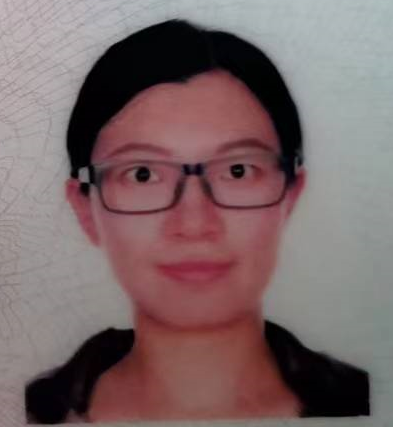}}]{Ling Yin}
received the B.S. degree in software engineering from East China Normal University, China, in 2008, and the Ph.D. degree in Computer technology from East China Normal University, China, in 2016. She is currently a lecturer with the School of Electronic and Electrical Engineering, Shanghai University of Engineering Science, China. Her research interest includes deep learning, time series analysis, and software engineering with formal methods.
\end{IEEEbiography}

\begin{IEEEbiography}
[{\includegraphics[width=1in,height=1.25in,clip,keepaspectratio]{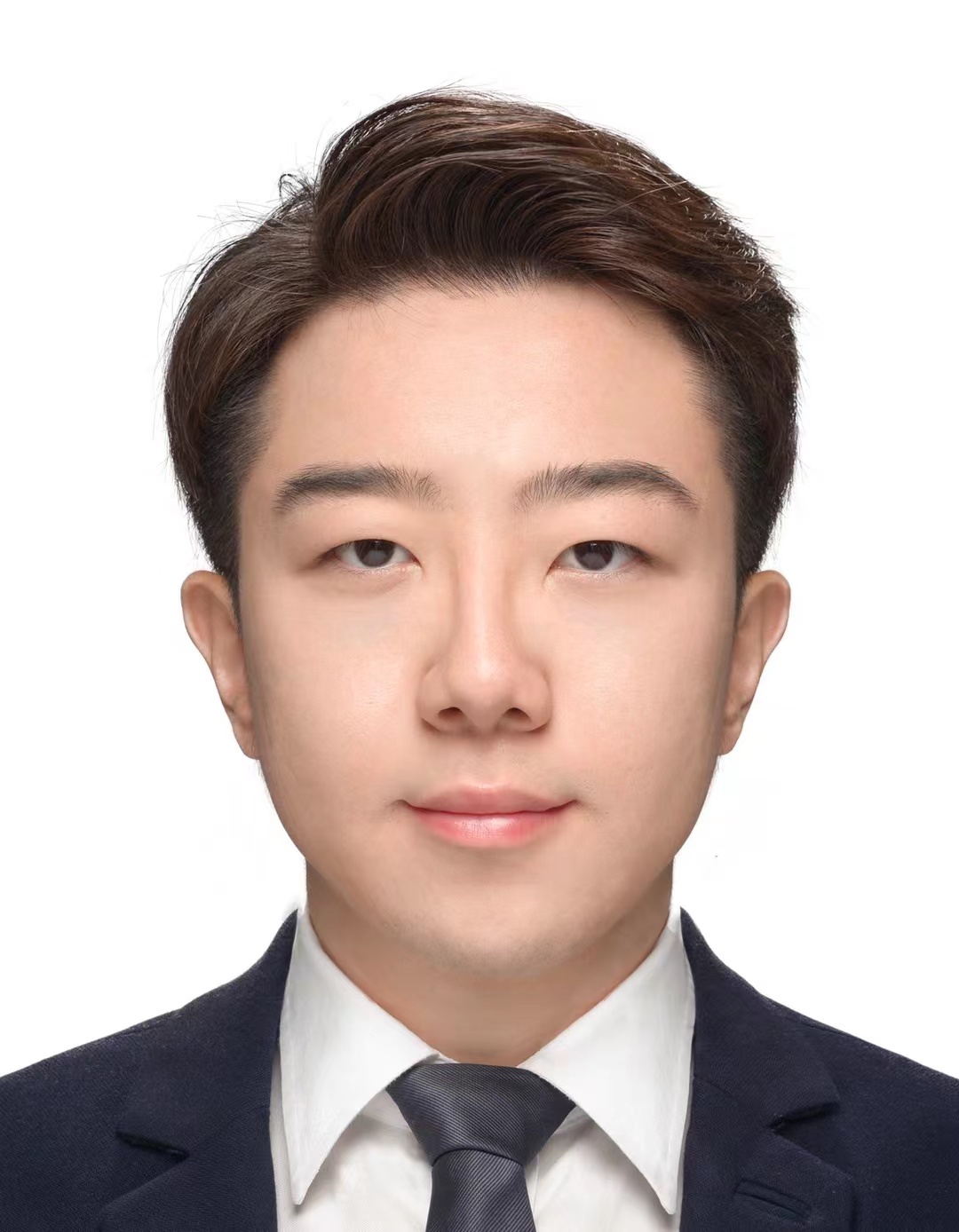}}]{Mingze Yang}
received the B.S.degree in Rolling Stock Engineering from Shanghai Institute of Technology, Shanghai, China, in 2019. He is currently a M.S. student with the department of electrical and electronic engineering of Shanghai University of Engineering Science, Shanghai, China. His research interests include deep learning, target recognition and wireless sensing.
\end{IEEEbiography}

\begin{IEEEbiography}
[{\includegraphics[width=1in,height=1.25in,clip,keepaspectratio]{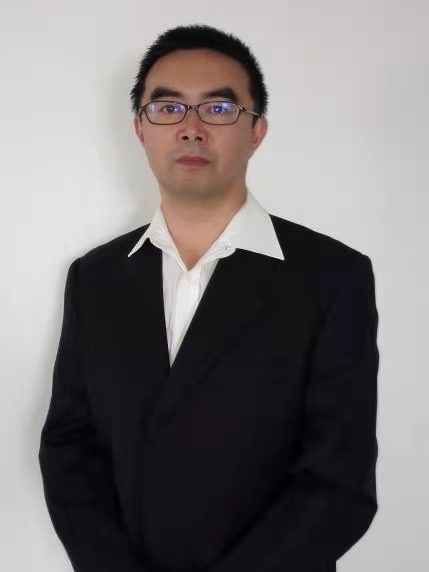}}]{Fei Wu}
received the B.S. degree, the M.S. degree and the Ph.D. degree in Computer Science from National University of Defense Technology in 1990, 1993 and 1998. He was a Post-Doctoral Research with Nankai University, China. He is currently a full professor with the School of Electronic and Electrical Engineering, Shanghai University of Engineering Science, China. His research interests include intelligent information processing, positioning technology and machine learning.
\end{IEEEbiography}

\begin{IEEEbiography}
[{\includegraphics[width=1in,height=1.25in,clip,keepaspectratio]{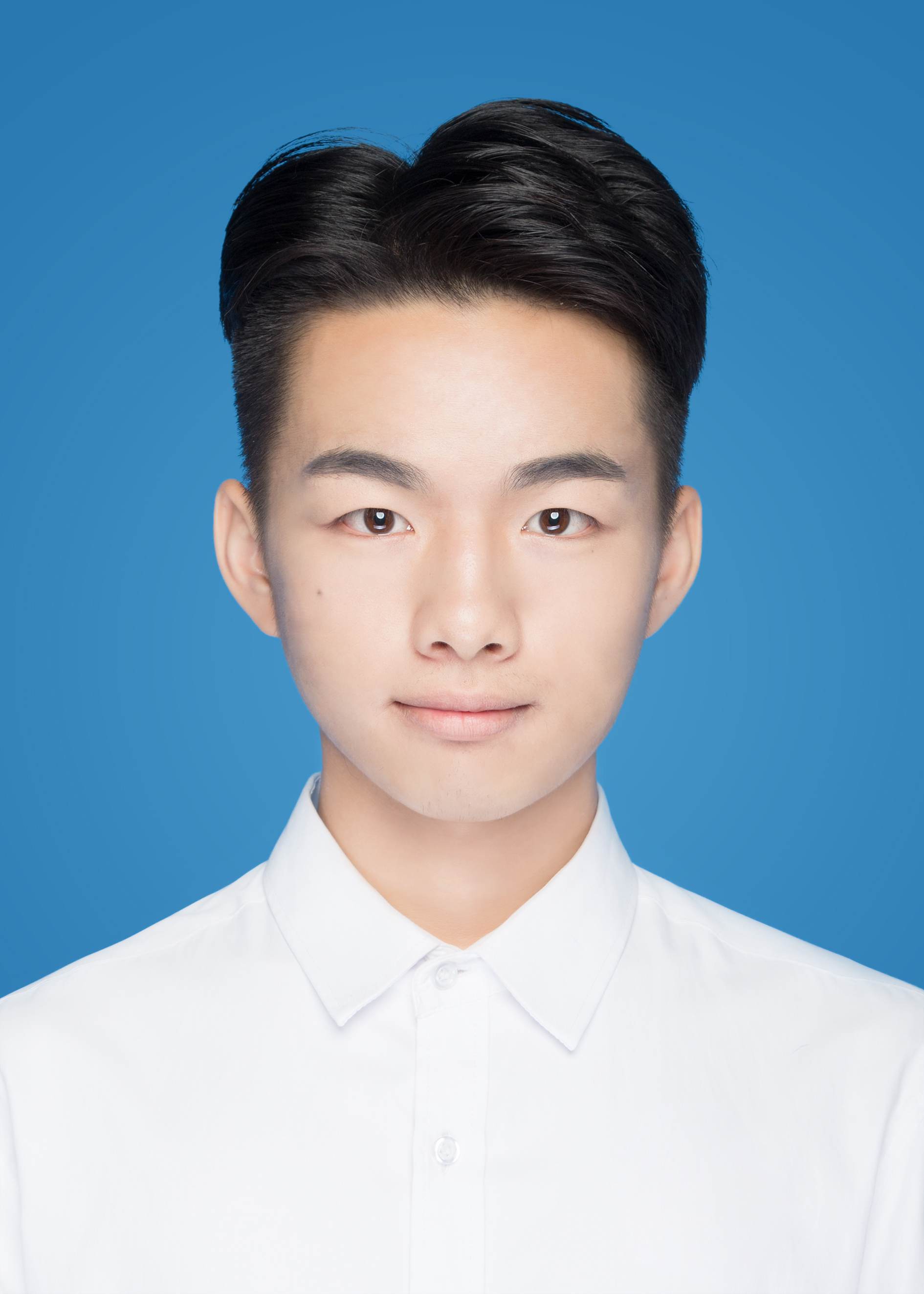}}]{Yuncheng Yang}
received the B.S.degree in mechanical engineering from Henan University of Science and Technology, Luoyang, China, in 2018. He is currently a M.S.student with the department of electrical and electronic engineering of Shanghai University of Engineering Science, Shanghai, China. His research interests include deep learning, indoor positioning and wireless sensing.
\end{IEEEbiography}

\begin{IEEEbiography}
[{\includegraphics[width=1in,height=1.25in,clip,keepaspectratio]{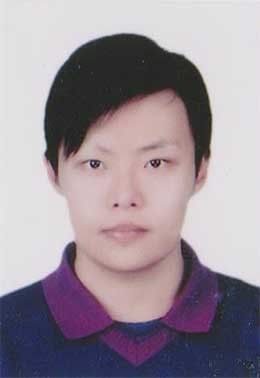}}]{Wenbo Hu}
 received the Ph.D. degree in Geography from Université Grenoble Alpes, Grenoble, France, in 2019. He is currently the post-doctor in School of Communication and Information Engineering, Shanghai University. From 2019 to 2020, he was a postdoctoral researcher with the Laboratoire PACTE, UMR 5194 CNRS, France. His research interest include behavioral geography, spatial modeling and public policing based on big data, deep learning and machine learning.
\end{IEEEbiography}

\vfill

\end{document}